\documentclass[5p]{elsarticle}

\usepackage{url}

\journal{Robotics and Autonomous Systems}

\makeatletter
\gdef\emailauthor#1#2{\stepcounter{ead}%
     \g@addto@macro\@elseads{\raggedright%
      \let\corref\@gobble
      \eadsep\texttt{#1}\def\eadsep{\unskip,\space}}%
}
\def\urlauthor#1#2{\g@addto@macro\@elsuads{\let\corref\@gobble%
    \raggedright\eadsep\texttt{#1}%
    \def\eadsep{\unskip,\space}}%
}
\makeatother

\usepackage{amsmath}
\usepackage{mathtools}
\usepackage{amssymb, bm}
\usepackage{amsfonts}
\usepackage{amsthm}
\usepackage{booktabs}
\usepackage{graphicx}
\usepackage{epstopdf}
\usepackage{caption,subcaption}
\usepackage{color}
\usepackage[dvipsnames]{xcolor}
\usepackage[ruled,vlined]{algorithm2e}
\usepackage{balance}
\allowdisplaybreaks
\usepackage{import}

\usepackage[usestackEOL]{stackengine}
\setstackgap{S}{1.5pt}
\usepackage{hhline}
\usepackage{multirow}
\definecolor{redc}{rgb}{1.0, 0, 0}

\usepackage{array}
\newcolumntype{C}[1]{>{\centering\arraybackslash}p{#1}}

\theoremstyle{definition}
\newtheorem{definition}{Definition}[]

\graphicspath{{figures/}}

\usepackage[labelfont=bf,justification=raggedright,singlelinecheck=false]{caption}
\usepackage{todonotes} 

\begin{document}

\begin{frontmatter}

\title{Distributed Allocation and Scheduling of Tasks with Cross-Schedule Dependencies for Heterogeneous Multi-Robot Teams}

\author[unizg]{Barbara Arbanas\corref{mycorrespondingauthor}}
\cortext[mycorrespondingauthor]{Corresponding author}
\ead{barbara.arbanas@fer.hr}
\ead[url]{www.larics.fer.hr}

\author[unizg]{Tamara Petrovi{\'{c}}}
\author[unizg]{Matko Orsag}
\author[unise]{J. Ramiro Mart{\'{i}}nez-de-Dios}
\author[unizg]{Stjepan Bogdan}

\address[unizg]{University of Zagreb, Faculty of Electrical Engineering and Computing,\\Unska 3, 10000 Zagreb, Croatia}
\address[unise]{Universidad de Sevilla, GRVC Robotics Lab Sevilla,\\Camino de los Descubrimientos s/n, 41092 Sevilla, Spain}

\begin{abstract}
To enable safe and efficient use of multi-robot systems in everyday life, a robust and fast method for coordinating their actions must be developed. In this paper, we present a distributed task allocation and scheduling algorithm for missions where the tasks of different robots are tightly coupled with temporal and precedence constraints. The approach is based on representing the problem as a variant of the vehicle routing problem, and the solution is found using a distributed metaheuristic algorithm based on evolutionary computation (CBM-pop). Such an approach allows a fast and near-optimal allocation and can therefore be used for online replanning in case of task changes. Simulation results show that the approach has better computational speed and scalability without loss of optimality compared to the state-of-the-art distributed methods. An application of the planning procedure to a practical use case of a greenhouse maintained by a multi-robot system is given.
\end{abstract}


\end{frontmatter}


\section{Introduction}
\label{sec:intro}

Research in the field of cooperative multi-robot systems (MRS) has received much attention among scientists in recent years, as it offers new possibilities compared to the use of a single robot. A multi-robot system can be homogeneous or heterogeneous \cite{Ismail2019}. In homogeneous robot teams, the capabilities of the individual robots are identical (the physical structures do not have to be the same) \cite{Dahl2009,Wawerla2010}. In heterogeneous robot teams, the capabilities of robots are different, allowing them to specialize in certain tasks \cite{Bicchi2008,Pei2012}. The main challenge is to provide a robust and intelligent control system so that the agents can communicate with each other and coordinate their tasks to accomplish an upcoming mission. Therefore, developing a robust control architecture, communication, and mission planning are the main problems discussed and solved in the literature. In this paper, we discuss the problems of task allocation (the question of \emph{who does what?}) and task scheduling (the question of \emph{how to arrange tasks in time?}) of multi-robot system missions, often summarized under the common term \emph{mission (task) planning}.

From the control architecture perspective, MRS can be divided into two groups -- centralized and distributed \cite{Ismail2019}. In centralized architectures, there is a central control agent that has global information and is in constant communication with the entire team to coordinate tasks and gather newly acquired information about the mission and the state of the world. Due to the global view of the mission, a centralized approach to task planning can produce optimal or near-optimal plans \cite{Zhang2015,Gombolay2018,Garca2013}. While centralized approaches may be appropriate for many applications, distributed architectures typically exhibit better reliability, flexibility, adaptability, and robustness. This makes them particularly suitable for field operations in dynamic environments where a larger number of robots with limited resources are deployed, even if the solutions they provide are often suboptimal.

The main advantages of distributed approaches in the literature are fast computation and scalability, which allow the system to respond on-the-fly to external disturbances such as mission changes and robot failures. Some of the best known distributed solutions to the mission planning problem are auction and market-based approaches \cite{Capitan2012,Choi2009,Nunes2017}. These generally fall into the domain of task allocation, where robots use bidding mechanisms for a set of simple tasks to allocate tasks among themselves. However, the partial ordering between tasks and tight coupling, which are the basis for our cooperative missions, are rarely considered. These constraints concern a class of problems \cite{korsahTaxonomy} where an agent's effective utility for a task depends on other tasks assigned to that agent (In-Schedule Dependencies, ID) or on the schedules of other agents (Cross-Schedule Dependencies, XD). More recently, the authors in \cite{Nunes2017,Mitiche2019} have addressed the problem of precedence constraints in iterative auctions. These algorithms can be run offline or online, and they solve the same class of problems as those considered in this work.

The tasks we model in this paper fall into the class of problems XD[ST-SR-TA] defined in the taxonomy in \cite{korsahTaxonomy}. These task types require execution by a single robot (SR, single-robot tasks), and robots are allowed to execute only one task at a time (ST, single-task robots). The task allocation and scheduling procedure considers both current and future assignments (TA, time-extended assignment). In terms of complexity, these tasks involve cross-schedule dependencies (XD), where various constraints relate tasks from plans of different robots.

On the other side of the spectrum, various optimization-based methods attempt to solve the task planning problem. They range from exact offline solutions \cite{Saribatur2014} to heuristic approaches such as evolutionary computation and other AI optimization methods \cite{Schillinger2018,Omidshafiei2017}. In the former, a broader class of problems is considered than in this work (XD[ST-MR-TA], MR stands for multi-robot tasks). The problem is modeled as an instance of a Mixed-Integer Linear Programming (MILP) problem and solved using offline solvers or well-known optimization methods. Although optimal, the method involves high computational burden and lacks reactivity in dynamic environments.

The decentralized framework for multi-robot coordination is proposed in our previous works \cite{Arbanas2018,Krizmancic2020}, but these works focus on coordination mechanisms, and only simple algorithms for task allocation and scheduling are given. In \cite{Arbanas2018}, tasks are assigned to robots using a greedy function that allots each task to the robot with the best score for that task, ignoring the effects on other task assignments. In recent work \cite{Krizmancic2020}, we extended the assignment function to a simple market-based task allocation scheme that aims to minimize the total mission duration.

\subsection*{Motivation and methodology}

In this paper, we propose a solution for distributed task allocation and scheduling problem in heterogeneous multi-robot teams for problems within the previously defined class of XD[ST-SR-TA]. The cross-schedule dependencies we consider are precedence constraints and transitional dependencies. The novelty of our solution is based on the fact that by representing the problem as a variant of the Vehicle Routing Problem (VRP), we can define a \emph{generic model of task planning problems} that can be applied to problems from different domains of multi-robot and multi-agent systems. The proposed solution acts as a domain-agnostic planner of problems that adhere to the specified model. Another advantage of the proposed modeling is that it exposes the task planning problems to a wide range of optimization techniques already available for VRP, thus advancing the state of the art in task planning.

Similar modeling was proposed in \cite{Korsah2011}, where the problem of task planning refers to the Dial-a-Ride Problem (DARP), a variant of VRP with pickup and delivery. To solve the problem, the authors use a centralized bounded optimal branch-and-price algorithm. In our approach, we employ multi-objective optimization with a form of distributed genetic algorithm using mimetism and knowledge sharing. This approach, which uses distributed evolutionary computation methods, can quickly generate near-optimal solutions, and thus, work online while achieving good scalability properties.

To empirically validate the proposed algorithm, we first evaluate our distributed solution on benchmark Multi-Depot Vehicle Routing Problem (MDVRP) dataset and its solutions available in the VRP-related literature. However, these benchmark problems do not cover all complexities of the considered \emph{task planning problems}, such as precedence constraints. Therefore, we generate a generic set of multi-robot task planning problems and compare our solution on them against the following techniques:

\begin{enumerate}
    \item Gurobi Optimizer \cite{gurobi}. Gurobi is a centralized solution that uses exact mathematical methods to solve MILP problems. Gurobi provides an optimality gap for each solution and can therefore serve as a measure of the optimality of the proposed solution. 
    \item State-of-the-art distributed auction-based approach with a single central agent acting as an auctioneer \cite{Nunes2017}. We slightly adapt the method to suit our problem class.
\end{enumerate}

We chose the Gurobi optimizer to measure the optimality of the solutions because it provides an exact mathematical approach to solving the problem. However, it can only generate solutions for the simpler problems in the dataset because its computational complexity increases rapidly with the dimensionality of the problem (larger number of tasks). From a very limited pool of distributed approaches in the literature, we selected the auction-based method \cite{Nunes2017} as a direct counterpart to our method. These two methods serve as benchmarks against which we evaluate the performance of our proposed approach.

\subsection*{Original contributions of the paper}
The first major contribution of this work is \emph{the unified task planning model} (allocation and scheduling) based on a VRP paradigm. It combines problems of task planning with the well-studied MDVRP model, which generalizes the problem and makes it domain-agnostic. This opens the door for the use of various optimization techniques available in the field of vehicle routing, and has the potential to yield many compelling solutions for task planning in the future.

The next contribution is the formulation of a distributed metaheuristic for the defined problem based on the Coalition-Based Metaheuristic (CBM) paradigm (CBM-pop\footnote{\url{https://github.com/barbara0811/cbm_pop_mdvrp_optimization}}). The algorithm, originally developed for a basic VRP problem \cite{Meignan2009}, is extended to a case of MDVRP with heterogeneous fleet under capacity and precedence constraints. In addition to adapting to a different problem, we have also contributed to the search procedure itself by improving the way solutions are handled. Instead of keeping only one solution, our algorithm works with a population of solutions, which makes it more efficient in finding optima.

Finally, we have created an open benchmark dataset repository for planning for tasks of class XD[ST-SR-TA] \cite{BenchmarkRepo}. The dataset consists of generic, domain-agnostic, randomly generated sets of task planning problems and provides an unbiased set of examples for evaluating planning algorithms. A similar dataset was created in \cite{Mitiche2019} for problems with a fixed number of tasks ($100$) and up to $7$ robots. Our proposed benchmark dataset includes problems with $2^n, 2 \leq n \leq 10, n \in \mathbb{N}$ tasks and $8$ robots, providing a large-scale test of the efficiency and scalability of task planning algorithms.

Through extensive testing of the proposed algorithm on the newly developed dataset against existing task planning methods, we have shown that our method is state-of-the-art in terms of optimality. The advantage of our approach is better computational speed and scalability, which is essential for any online operation of multi-robot systems. The method is then applied to a case of multi-robot collaboration in a robotized greenhouse, as presented in the SpECULARIA project \cite{specularia}.


These contributions are disseminated in the paper as follows. In the next section, we outline the approaches to MDVRP problems and reason on the solution we have chosen for our problem. In Section \ref{sec:problem_statement}, we formally define the task planning problem. In Section \ref{sec:VRP}, we give VRP preliminaries used in task modeling, followed by the model definition that relates the concepts and definitions established in the preceding two sections to form a unified model of task planning based on the paradigm of capacitated MDVRP with heterogeneous fleet. In Section \ref{sec:solution_approach}, we present our distributed approach to solving the defined MDVRP. Section \ref{sec:results} contains a thorough evaluation of the results and a discussion. In the same section, we show how the proposed approach can be applied to the described use-case example. At the end of the paper, we give a conclusion and future work.

\section{MDVRP Solution Approaches}

The problems considered here (XD[ST-SR-TA]) fit well into the paradigm of the Multi-Depot Vehicle Routing Problem (MDVRP). We regard a MDVRP variant with heterogeneous fleet and capacity constraints to incorporate all relevant aspects of task planning. MDVRP is a VRP variation concerned with problems of servicing customers from several depots with a designated fleet of vehicles. It is a classic example of an NP-hard \cite{Garey1990} combinatorial optimization problem. Even for relatively small problem sizes, solving MDVRP to optimality is challenging. Given the rapid combinatorial explosion, it quickly becomes impossible to obtain optimal solutions for this type of problems.

The authors in \cite{Laporte1984} were the first to report on optimal solutions for problem sizes up to $50$ customers using a branch-and-bound method. The method was soon improved for asymmetric MDVRPs by \cite{Laporte1988}, who first transformed the problem into an equivalent constraint assignment problem and then applied a branch-and-bound technique to problem instances with up to 80 customers and three depots. More recently, \cite{Baldacci2009} developed an exact method for solving the Heterogeneous Vehicle Routing Problem (HVRP) that can be used to solve several variants of VRP, including MDVRP. 
The authors present computational results for MDVRP instances with up to $200$ customers and $2$-$5$ depots. The authors in \cite{Contardo2014} present a new exact algorithm for MDVRP under capacity and route length constraints. The model is defined using a vehicle-flow and a set-partitioning formulation, which are used in different phases of the algorithm. Their method is based on variable fixing, column-and-cut generation, and column enumeration. In their work, optimality has been proven for the first time for some benchmarking instances.

On the other hand, several heuristic methods have been proposed for MDVRP problems. These approaches seek approximate solutions in polynomial time instead of computationally expensive exact solutions. The work in \cite{MontoyaTorres2015} revealed that most researchers tend to solve the MDVRP by heuristics or metaheuristics. In the field of global optimization, \emph{metaheuristics} are stochastic search algorithms specified as generic algorithm frameworks that use rules or heuristics applicable to different problems to accelerate their convergence to near-optimal solutions \cite{Basset2018}. In general, metaheuristics emulate processes and behaviors inspired by mechanisms found in nature, such as evolution. One of the most popular metaheuristic algorithms is Genetic Algorithm (GA), and several variants have been proposed to solve MDVRP \cite{Berman2008, Gesu2005, Mahmud2019}. One of the main advantages of the GA approaches is their simplicity and fast computation, which allows them to quickly explore a large solution space and effectively generate suboptimal solutions. Other heuristic approaches include Ant Colony Optimization (ACO) algorithm \cite{Narasimha2013, Zhang2019}, Tabu search \cite{Escobar2014}, and various hybrid methods that combine multiple techniques \cite{Luo2014, Mirabi2010, Haerani2017, Stodola2018, deOliveira2016, HesamSadati2020}.

The metaheuristics are closely related to multi-agent model, as both can exploit the social metaphor and the self-organization paradigm. Recently, the field of Distributed Artificial Intelligence (DAI) has grown, including multi-agent systems that solve difficult combinatorial problems. The multi-agent concepts can be easily applied to metaheuristics, especially population-based, hybrid, and distributed metaheuristics. The advantages of the distributed approach are the apparent increase in computational power due to the simultaneous execution of multiple tasks and the increase in the robustness or efficiency of the search, which is fostered by cooperation and interaction between agents. Our proposed algorithm is inspired by the Coalition-Based Metaheuristic approach \cite{Meignan2009}, which uses previously defined principles of DAI to solve the VRP problem.
\section{Problem Statement}
\label{sec:problem_statement}

In this paper we consider three mathematical formulations. First, we formally define the task planning problem in this section, followed by the mathematical definition of VRP problems, and finally we propose a unified model that combines the two paradigms. The solution to the task planning problem proposed in this paper is based on the unified mathematical model.

We consider a problem where a team of heterogeneous robots $R = \{1,\ldots,m\}$ is available to perform a collection of simple single-agent tasks (\emph{actions}) $A = \{1,\ldots,n\}$. One or more robots can perform each $a \in A$, and we specify the set of actions that robot $i$ can perform as $A_i$. Redundancy is possible, and in general $A_i \cap A_j \neq \emptyset$, $i \neq j,$ and $i, j \in R$. 

In addition, we define \emph{precedence constraints} on the set of actions. If the action $a \in A$ must be completed before the action $b \in A$ starts, we can specify a constraint between the two as $prec(a, b)$. This constraint forces $a^f < b^s$, where $a^f$ and $b^s$ indicate the times when the action $a$ finishes and $b$ starts.

In addition to strict precedence constraints, the task planning problem also involves \emph{transitional dependencies} between actions. The problem involves the transition time between two actions (\emph{setup time}), which includes all the operations required to go from the execution of one action to the start of processing the next action. This term represents an additional temporal dependency on the order of actions, since direct predecessors and successors of actions strongly influence the total duration of the mission and the associated costs (\emph{setup costs}).

Another naturally occurring constraint inherent in the physical system itself is the \emph{capacity constraint}, which is expressed in the limited battery resources of each robot. Each action requires a certain amount of energy to be performed, which depends on the physical properties of the robot and the current state of the system (i.e., robot position, battery status, and current payload). For a robot $i \in R$ with the capacity $Q_i$, we define the capacity constraint as $\sum_{a \in S_i} q_i(a) \leq Q_i$, where $S_i (S_i \subseteq A_i)$ is the set of actions assigned to the robot $i$ for execution, and $q_i(a)$ is the energy requirement of the action $a$ to be completed by robot $i$.

A solution to the described problem is a set of time-related actions (\emph{schedule}) for all robots that do not violate the specified constraints. Formally, the schedule $s_i$ for each robot $i \in R$ is defined as $s_i = \{(a, a^s, a^f)\ \forall a \in S_i\}$, where $S_i$ is the set of actions assigned to the robot $i$, and $a^s (a^f)$ are the start (finish) times of the action $a$.

In the evaluation procedure, each action $a \in A$ is assigned a couple $(d_a(i), c_a(i)), \forall i \in R$, where $d_a(i)$ stands for the duration of the action $a$ when performed by robot $i$ and $c_a(i)$ for the cost of the action. The robot estimates the duration and cost of a future task based on the current state of the system.

The planning procedure aims to find a solution that meets all constraints and maximizes the global reward of the system. The objective function can be chosen at will, depending on the system requirements. In the case of multi-objective optimization, optimal decisions must be made in the presence of tradeoffs between two or more objectives that may be conflicting. In our solution, we have chosen a multi-objective function based on the principle of Pareto optimality (\cite{CHANG2015}), which looks for solutions where no preference criterion can be improved without degrading at least one criterion. The optimization method proposed in this paper is guided by an algorithm based on evolutionary computation and distributed artificial intelligence, where solutions are found by a distributed search in a large solution space. The algorithm is developed on the premise that the communication network of robots forms a connected graph. Interruptions and delays in the communication channel are not the focus of this work. More details on the solution approach follow in Section \ref{sec:solution_approach}.
\section{Preliminaries -- Capacitated Vehicle Routing Problem}
\label{sec:VRP}

In our approach, we model the task allocation and scheduling problem as a variant of VRP, capacitated MDVRP with heterogeneous fleet \cite{Salhi2014}, hereafter referred to as C-MDVRP. In this section, we outline the fundamental concepts of the VRP paradigm and present a single-depot variant of the capacitated VRP problem (CVRP), which is extended in the next section for the case of multiple depots.

CVRP is the most widely used VRP variant due to its numerous practical applications in transportation, distribution, and logistics. Essentially, CVRP is a problem where vehicles with limited payloads need to pick up or deliver items at different locations. The items have a quantity, such as weight or volume, and the vehicles have a maximum capacity that they can carry. The problem is to pick up or deliver the items at the lowest cost without exceeding the vehicle capacity.

The basic VRP \cite{toth2015VRP} regards a set of nodes $N = \{1,\ldots,n\}$ representing $n$ \emph{customers} at different locations and a central \emph{depot} (warehouse), which is usually denoted by $0$. Customers are served from one depot by a homogeneous and limited fleet of vehicles. A vehicle serving a customer subset $S \subseteq N$ starts at the depot, travels once to each customer in $S$, and finally returns to the depot. Each pair of locations $(i,j)$, where $i,j \in N \cup \{0\}$, and $i \neq j$, is associated with a \emph{travel cost} $c_{ij}$ that is symmetric, $c_{ij} = c_{ji}$.

In the CVRP, each customer is assigned a \emph{demand} $q_i, i \in N$ that corresponds to the quantity (e.g., weight or volume) of goods to be delivered from the depot to the customer. There is a set of \emph{vehicles}, $K = \{1,\ldots,m\}$, with capacity $Q >0$, operating at identical cost. In the case of a heterogeneous fleet, the capacity $Q$ is specifically defined for each vehicle (or type of vehicle).

A \emph{route} is a sequence $r = (i_0,i_1,\ldots,i_s,i_{s+1})$ with $i_0 = i_{s+1} = 0$, and $S = \{i_1,\ldots,i_s\} \subseteq N$ is the set of visited customers. The route $r$ has cost $c(r) = \sum_{p=0}^s{c_{i_p i_{p+1}}}$. A route is considered feasible if the capacity constraint $q(S) \coloneqq \sum_{i\in S}{q_i} \leq Q$ holds and no customer is visited more than once, $i_j \neq i_k$ for all $1 \leq j < k \leq s$. In this case, the set $S \subseteq N$ is considered a \emph{feasible cluster}.

A solution of a CVRP consists of $m = |K|$ feasible routes, one for each vehicle $k \in K$. $|K|$ represents the cardinality of the set $K$. Therefore, the routes $r_1,r_2,\ldots,r_m$ corresponding to the specific clusters $S_1,S_2,\ldots,S_m$ represent a feasible solution of the CVRP if all routes are feasible and the clusters form a partition of $N$.

The given model can be represented by an undirected or directed graph. Let $V = \{0\} \cup N$ be the set of vertices (or nodes). In the symmetric case, i.e., if the cost of moving between $i$ and $j$ does not depend on the direction, the underlying graph $G = (V,E)$ is complete and undirected with edge set $E = \{e=(i,j)=(j,i): i,j\in V, i \neq j\}$ and edge cost $c_{ij}$ for $(i,j) \in E$. Otherwise, if at least one pair of vertices $i,j \in V$ has asymmetric cost $c_{ij} \neq c_{ji}$, then the underlying graph is a complete digraph. We are concerned with the former.

\begin{definition}{\emph{ CVRP model formulation.}} One of the most common mathematical representations of the VRP model is the MILP formulation \cite{toth2015VRP}. The binary decision variable $x_{ijk}$ is defined to indicate whether the vehicle $k, k \in K$ traverses an edge $(i,j) \in E$ in a given solution. Therefore, the integer linear programming model for the CVRP can be considered as written:

\begin{subequations}
\renewcommand{\theequation}{\theparentequation.\arabic{equation}}
\begin{alignat}{3}
    & \text{(CVRP)} \qquad \qquad \qquad 
       min\ \sum_{k \in K} \sum_{(i,j)\in E} && c_{ij}x_{ijk} \label{eq:CVRP-obj}\\
    \intertext{Subject to}
    &\sum_{k \in K} \sum_{i \in V, i \neq j} x_{ijk} = 1, && \forall j \in V \setminus \{0\}, \label{eq:CVRP-degree}\\
    &\sum_{j \in V \setminus \{0\}} x_{0jk} = 1, && \forall k \in K, \label{eq:CVRP-depot}\\
    &\sum_{i \in V, i \neq j} x_{ijk} = \sum_{i \in V} x_{jik}, && \forall j \in V, k \in K, \label{eq:CVRP-flow}\\
    &\sum_{i \in V} \sum_{j \in V \setminus \{0\}, j \neq i} q_j x_{ijk} \leq Q, && \forall k \in K, \label{eq:CVRP-capacity}\\
    &\sum_{k \in K} \sum_{i\in S} \sum_{j\in S, j \neq i} x_{ijk} \leq |S| - 1, \quad && \forall S \subseteq N, \label{eq:CVRP-subtour}\\
    &x_{ijk} \in \{0,1\}, && \forall k \in K, (i,j) \in E. \label{eq:CVRP-x-domain}
\end{alignat}
\end{subequations}
The objective function (\ref{eq:CVRP-obj}) minimizes the total travel cost. The constraints (\ref{eq:CVRP-degree}) are the degree constraints that ensure that exactly one vehicle visits each customer. The constraints (\ref{eq:CVRP-depot}) and (\ref{eq:CVRP-flow}) guarantee that each vehicle leaves the depot only once, and that the number of vehicles arriving at each customer and returning to the depot is equal to the number of vehicles departing from that node. Capacity constraints are expressed in (\ref{eq:CVRP-capacity}), and ensure that the sum of the demands of the customers visited on a route is less than or equal to the capacity of the vehicle providing the service. The sub-tour elimination constraints (\ref{eq:CVRP-subtour}) ensure that the solution does not contain cycles disconnected from the depot. The constraints (\ref{eq:CVRP-x-domain}) specify the domains of the variables. This model is known as a three-index vehicle flow formulation.
\end{definition}
\section{Task Planning Model Formulation}
\label{sec:vrp_model}
 
In the previous section, we introduced basic VRP concepts for the single-depot capacitated VRP model, and now we extend the modeling to multiple depots with heterogeneous fleet (C-MDVRP model), and relate it to the task planning paradigm. In this section, we first describe the elements of the C-MDVRP through the lens of task planning, equating all the building blocks of the VRP-based model with a concept of task planning as defined in the previous two sections. Then, we define the complete unified mathematical model in terms of a MILP.

\subsection{Task planning as C-MDVRP}

\begin{figure*}[htb]
    \centering
    \includegraphics[width=0.7\linewidth,trim=80pt 40pt 80pt 80pt,clip]{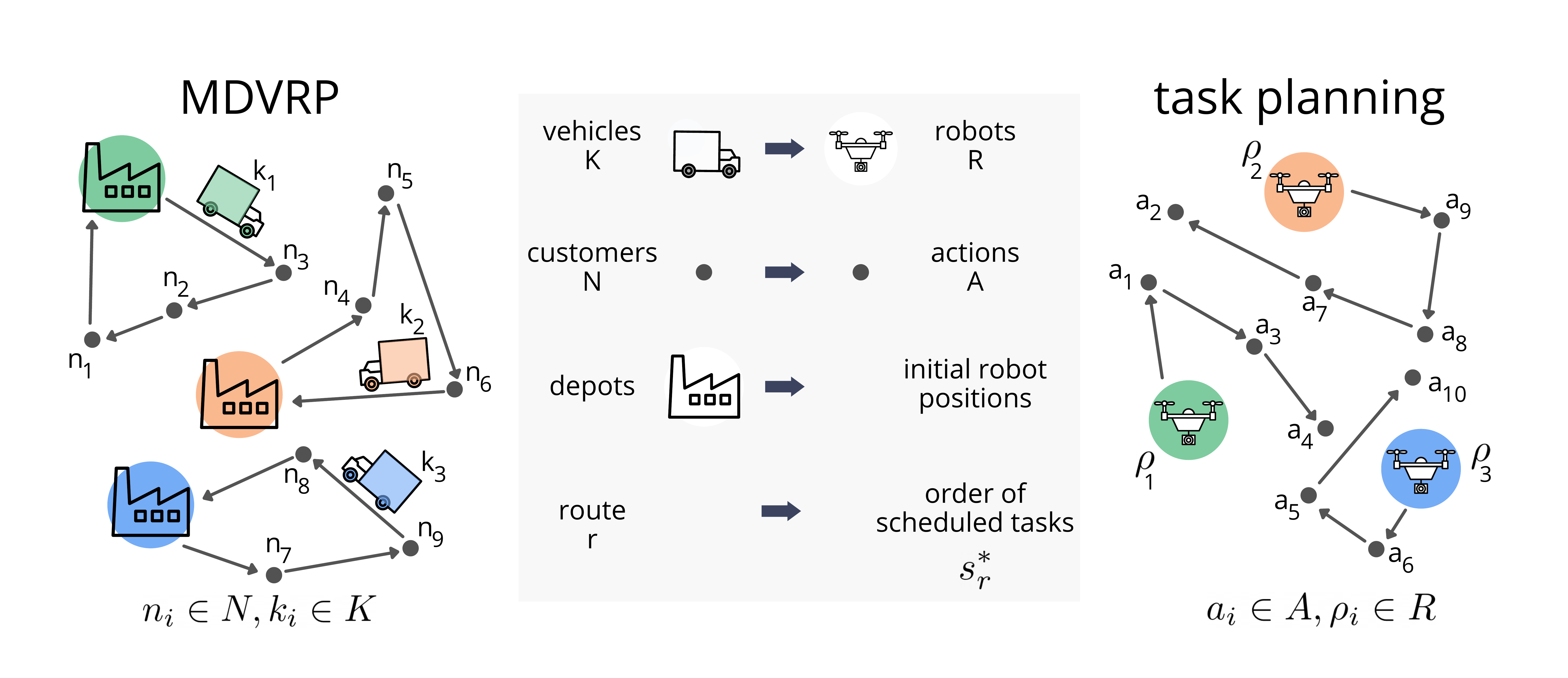}
    \caption{An illustration of the relationship of the C-MDVRP model to the task planning paradigm. The diagram shows the direct relation between concepts in VRP and task planning problems, where depots represent initial robot positions, vehicles stand for robots, customers refer to actions, and vehicle routes represent final robot schedules.}
    \label{fig:VRP_illustration}
\end{figure*}

In relating the task planning problem to the C-MDVRP model, we associate the fundamental VRP concepts directly to the task planning paradigm. As shown in Figure \ref{fig:VRP_illustration}, the idea of a depot in VRP problems is directly related to the initial position of the robot, and the vehicle in VRP stands for a robot itself. Next, the concept of customer and customer demand is applied to actions in task planning and the energy requirement of each action, respectively. Consequently, routes as solutions to VRP problems represent the sequence of actions in the final robot schedules in the task planning model. Based on the metaphor thus specified, we describe the mathematical model of task planning problems defined as a VRP problem variant.

\begin{table*}[htb]
\centering
\setlength{\tabcolsep}{0.5pc}
\newlength{\digitwidth} \settowidth{\digitwidth}{\rm 0}
\catcode`?=\active \def?{\kern\digitwidth}
    \caption{Task planning elements as C-MDVRP.}
    \label{tbl:task_planning_vrp}
    \resizebox{0.8\textwidth}{!}{\begin{tabular}{*{3}{l}}
        \hline
         Task Planning & C-MDVRP & Unified Model Formulation \\
         \hline
         actions, $A$          &    customers, $N$    &   $A$, set of actions \\
         robots, $R$           &    vehicles, $K$     &   $R$, set of robots \\
         schedule, $s$         &    route, $r$        &   $s^*_r = (i_0(r),i_1,\ldots,i_{|S_r|}), i \in A, r\in R$ \\
         start of the action, $a_i^s$          &   start of the service at node & $\omega_{ir}, i \in A, r \in R$ \\
         action duration, $d_i$          &   service duration    & $\sigma_{ir}, i \in A, r \in R$ \\
         end of the action, $a_i^f$            &   end of the service at node  & $\omega_{ir} + \sigma_{ir}, i \in A, r \in R$ \\
         initial position of the robot &    depot              &   $i_0(r), r \in R$, initial zero-cost action \\ 
         action setup cost             &   travel cost         &   $c_{ijr}, i,j \in \{0\} \cup A, r \in R$\\
         setup time             &   travel time         &   $t_{ijr}, i,j \in \{0\} \cup A, r \in R$\\
         energy requirement of the action     &   customer demand     &   $q_{ir}, i \in A, r \in R$\\
         robot energy capacity         &   vehicle capacity    &   $Q_r, r \in R$ \\
         \hline
    \end{tabular}}    
\end{table*}

In the modeling of task planning as a VRP variant, we use the term \emph{customer} for the simple single-robot tasks, \emph{actions}. We designate the set $N = \{1,\ldots,n\} = A$, where $A$ is the set of actions. This definition includes only actionable elements from the robot's task structure, and all tasks must be decomposed down to the action level where vehicle routing is optimized.

In the task planning paradigm, we directly equate the concept of a \emph{vehicle} with \emph{robots}, $K = \{1,\ldots,m\} = R$. Typically, the VRP paradigm requires vehicles to start at the depot, serve assigned customers along the route, and return to the depot. In our modeling, we equate the term \emph{depot} with the \emph{initial location} of the robot. Unlike the vehicles in typical VRP problems, the robot is not required to return to the starting point. Such a variant of VRP is referred to in the literature as \emph{open VRP} \cite{Li2007openVRP}.

The proposed model distinguishes between two different cost variants. On the one hand, the cost of transitioning between two tasks, \emph{setup cost}, is directly related to the \emph{travel cost} of the C-MDVRP. This cost can include any movement between locations and the possible setup cost between two tasks and is defined as $c_{ijr}, i,j \in \{0\} \cup A, i \neq j, r \in R$. Closely related to the setup cost is the \emph{setup time}, $t_{ijr}, i,j \in \{0\} \cup A, i \neq j, r \in R$, which defines the duration of the robot's setup between two tasks. Next, the \emph{energy requirement} of the specific action $i \in A$ is defined as $q_{ir}, r \in R$ and characterized as the customer \emph{demand} in C-MDVRP.

According to the specified modeling, the robot \emph{schedules} are constructed based on the \emph{routes} in the C-MDVRP solution. The parallel is apparent, since both constructs represent temporally ordered sequences of jobs. In the open VRP paradigm, a route is a sequence $r = (i_0,i_1,\ldots,i_s)$ with $i_0 = 0$, where $0$ denotes a depot node. In our modeling, we take the previously defined concept of a schedule, and map the order of tasks in a schedule to the arranged sequence of tasks $s^*_r = (i_0,i_1,\ldots,i_{|S_r|}), i_j \in S_r, j \in \{1, \ldots, |S_r|\}, r\in R$, where $i_0$ is a zero-cost task associated with the robot's starting location, and $S_r$ is a set of scheduled tasks. Temporal elements of the schedule are the task \emph{start time} and the task \emph{duration}, defined respectively as: $\omega_{ir} \coloneqq a_i^s, i \in S_r, r \in R$ and $\sigma_{ir} \coloneqq d_i = a_i^f - a_i^s, i \in S_r, r \in R$.

A schedule for a robot $r \in R$ is considered feasible if the capacity constraint $q_r(S_r) \coloneqq \sum_{i\in S_r}{q_{ir}} \leq Q_r$ holds, and no task is scheduled more than once, $i_j \neq i_k$ for all $1 \leq j < k \leq |A|$. When it is necessary to perform repetitive tasks, each occurrence of the task should be treated as a separate entity.

All previously defined relations are summarized in the Table \ref{tbl:task_planning_vrp} and additionally illustrated in Figure \ref{fig:VRP_illustration}.

\subsection{Mathematical model formulation}

The mathematical representation of the unified task planning model as MILP is based on the graph structure. Let $V$ be the set of vertices (or nodes) consisting of two distinct sets of nodes, the action nodes $V_a = A = \{1,\ldots,n\}$ and the start position nodes $V_s = \{1,\ldots,l\}$, where $n$ and $l$ represent the number of available tasks and robot start positions, respectively. It holds that $V = V_a \cup V_s$ and $V_a \cap V_s = \emptyset$. The underlying graph $G = (V,E)$ is complete and directed with edge set $E = \{e=(i,j): i,j \in V, i \neq j, i$ and $j$ not both in $V_s\}$.

We model heterogeneity motivated by a mixed fleet variant of MDVRP, in which the set of customers available to the vehicle corresponds in the task planning formulation to a subset of actions $A_r$ that the robot $r \in R$ can perform. To model unavailable tasks $j \in A \setminus A_r$, we set $c_{ijr}$ to a sufficiently large number $M$ for all $(i,j) \in E$ related to the unreachable task $j$. Further aspects of heterogeneity of multi-robot systems are achieved by replacing the general coefficients of MDVRP with robot-specific ones, e.g., the capacity $Q$ by $Q_r$, the energy requirement $q_i$ by $q_{ir}$ for all $i \in V_a$, and the cost $c_{ij}$ by $c_{ijr}$ for all $(i,j) \in E$.

For the given precedence constraints $prec(a_i, a_j)$, $a_i, a_j \in A$ and assuming that tasks $a_i (a_j)$ correspond to nodes $i(j) \in V_a$, we represent the precedence constraint in VRP notation by adding the pair of tasks to the set of constrained tasks $\Pi =\{(i, j), i, j \in V_a\}$. The constraint on the tasks is then
\begin{equation}
    \centering
    \omega_i + \sigma_i \leq \omega_j, (i,j) \in \Pi,
\end{equation}
where $\omega_i$ and $\omega_j$ mark the beginning of the tasks $i$ and $j$ respectively, and $\sigma_i$ denotes the duration of task $i$. We also distinguish particular $\omega_{ir}$ as the start time of task $i$ when it is executed by robot $r \in R$, and the corresponding task duration $\sigma_{ir}$. The expressions for these cases are defined in the full model representation.

\begin{table}
    \centering
    \caption{Defined sets in the unified model.}
    \label{tbl:task_planning_sets}
    \begin{tabular}{ll}
         \hline
         Set & Definition \\
         \hline
         $R$        & set of robots \\ 
         $V$        & set of vertices (nodes) \\ 
         $V_s$      & set of robot initial locations \\ 
         $V_a = A$  & set of single-robot actions \\ 
         $E = V \times V$ & set of edges \\ 
         $R_i$      & set of robots starting from location $i, i \in V_s$ \\ 
         $\Pi$      & set precedence constraints \\ \hline
    \end{tabular}
\end{table}

\begin{table}
    \centering
    \caption{Defined variables and constants in the unified model.}
    \label{tbl:task_planning_variables}
    \begin{tabular}{clc}
         \hline
         Variable & Definition & Domain \\
         \hline
         $x_{ijr}$  & if robot $r$ traverses edge $(i,j) \in E$ & $\{0,1\}$ \\ 
         \multirow{2}{*}{$\omega_{ir}$}   & \multirow{2}{*}{\Shortunderstack[l]{start time of action $i \in V_a$\\ performed by $r \in R$}} & \multirow{2}{*}{$\mathbb{R}$} \\
         & & \\
         \hline
         Constant && \\
         \hline
         \multirow{2}{*}{$\sigma_{ir}$}       & \multirow{2}{*}{\Shortunderstack[l]{duration of action $i \in V_a$ when\\ performed by robot $r \in R$}} & \multirow{9}{*}{$\mathbb{R}$} \\ 
         & & \\
         \multirow{2}{*}{$t_{ijr}$}      & \multirow{2}{*}{\Shortunderstack[l]{setup time between actions $i$ and $j$,\\$\forall (i,j) \in E$ assigned to $r \in R$}} & \\ 
         & & \\
         \multirow{2}{*}{$c_{ijr}$}      & \multirow{2}{*}{\Shortunderstack[l]{setup cost between actions $i$ and $j$,\\$\forall (i,j) \in E$ assigned to $r \in R$}}  & \\ 
         & & \\
         \multirow{2}{*}{$q_{ir}$}       & \multirow{2}{*}{\Shortunderstack[l]{energy demand of action $i \in V_a$\\for robot $r \in R$}} & \\ 
         & & \\
         $Q_r$          & energy capacity of robot $r \in R$ & \\ \hline
    \end{tabular}
\end{table}

For convenience, all sets, variables, and constants of the model are summarized in Tables \ref{tbl:task_planning_sets} and \ref{tbl:task_planning_variables}.

\begin{definition}{\emph{Unified task planning model formulation.}} Based on the relations defined in this section, we define the unified task planning model built according to the C-MDVRP MILP formulation. The binary decision variable $x_{ijr}$ is defined to indicate whether the robot $r \in R$ traverses an edge $(i,j) \in E$ in a given solution. Then, the model is given as:
\begin{subequations}
\renewcommand{\theequation}{\theparentequation.\arabic{equation}}
\begin{align}
    &\text{min} \qquad \quad \begin{cases}
      \delta =  max(\sum_{r \in R} \sum_{i' \in V} x_{i'ir} (\omega_{ir} +  \sigma_{ir})) - \\ \qquad min(\sum_{r \in R} \sum_{j' \in V} x_{j'jr} \omega_{jr}) \\
      \gamma = \sum_{r \in R} \sum_{(i,j)\in E} x_{ijr} (c_{ijr}+ q_{jr})
    \end{cases} \label{eq:unified-model-obj-min}
\end{align}
\begin{alignat}{3}
    \intertext{Subject to}
    &\sum_{r \in R} \sum_{i \in V, i \neq j} x_{ijr} \leq 1, && \forall j \in V_a, \label{eq:unified-model-degree}\\
    &\sum_{j \in V_a} x_{ijr} \leq 1, && \forall i \in V_s, r \in R_i, \label{eq:unified-model-depot}\\
    &\sum_{(i,j)\in E} x_{ijr}(\omega_{ir} + \sigma_{ir} + t_{ijr} - \omega_{jr}) \leq 0, && \forall r \in R \label{eq:unified-model-sched}\\
    &\sum_{r \in R} \sum_{i' \in V} x_{i'ir} (\omega_{ir} + \sigma_{ir}) \leq \sum_{r \in R} \sum_{j' \in V} x_{j'jr} && \omega_{jr}, \forall (i,j) \in \Pi, \label{eq:unified-model-prec}\\
    &\sum_{i \in V} \sum_{j \in V \setminus \{0\}, j \neq i} q_{jr} x_{ijr} \leq Q_r, && \forall r \in R, \label{eq:unified-model-capacity}\\
    &\sum_{r \in R} \sum_{i\in S} \sum_{j\in S, j \neq i} x_{ijr} \leq |S| - 1, \quad && \forall S \subseteq V_a, \label{eq:unified-model-subtour}\\
    &x_{ijr} \in \{0,1\},   && \forall r \in R, (i,j) \in E, \label{eq:unified-model-x-domain}\\
    &\omega_{ir} \in \mathbb{R},   && \forall i \in V, r \in R. \label{eq:unified-model-omega-domain}
\end{alignat}
\end{subequations}
The objectives of the optimization problem are represented by Equations (\ref{eq:unified-model-obj-min}). The goal is to find a solution that minimizes the makespan $\delta$ (the difference between the latest action finish time in the whole mission and the earliest action start) and the cost $\gamma$. Actions $i'$ and $j'$ are possible direct predecessors of actions $i$ and $j$, respectively, in the schedules of (possibly different) robots. The sum $\sum_{i' \in V} x_{i'ir}$ is equal to $1$ for exactly one action $i'$ that precedes $i$ in the route of robot $r$. Constraints (\ref{eq:unified-model-degree}) require that each task is executed at most once. Equations (\ref{eq:unified-model-depot}) state that each robot may execute at most one schedule (only one edge starting at robot's initial position can be incorporated in the solution). Note that flow constraints represented in previously defined versions of VRP in Equations (\ref{eq:CVRP-flow}) are omitted here. This is because we do not restrict this model to provide closed-loop solutions, as we consider the open VRP formulation. Next, constraints (\ref{eq:unified-model-sched}) guarantee schedule feasibility with respect to considerations in the schedule of each robot. The $t_{ijr}$ represents the setup time between actions $i$ and $j$ for robot $r$. The constraints defined by (\ref{eq:unified-model-prec}) enforce precedence constraints. Equations (\ref{eq:unified-model-capacity}) ensure that the capacity constraints are met by the robots, while (\ref{eq:unified-model-subtour}) eliminate all possible sub-tours in the solution. Finally, variable domains are provided in (\ref{eq:unified-model-x-domain}) and (\ref{eq:unified-model-omega-domain}).
\end{definition}
\section{Solution Approach}
\label{sec:solution_approach}

In this section, we provide insight into the proposed algorithm inspired by the CBM paradigm to solve the problem defined in Section \ref{sec:vrp_model}. We briefly describe the motivating algorithm on which our approach is based, and then elaborate on the specifics of the implementation and the introduced novelty.

\subsection{The Coalition-Based Metaheuristic}

In CBM \cite{Meignan2009}, multiple agents organized in a coalition simultaneously explore the solution space, cooperate, and self-adapt to solve the given problem collectively. The novelty introduced in this algorithm was the use of basic DAI principles, reinforcement and mimetic learning, which not only allow agents to learn from their experiences and adapt their future behaviors accordingly, but also to share knowledge with other agents in the coalition. In addition to the learned behaviors, the agents also share the best solutions found, so that at the end of each iteration of the algorithm, the best global solution to the problem is obtained.

\begin{figure}[htb]
    \centering
    \includegraphics[width=0.9\linewidth]{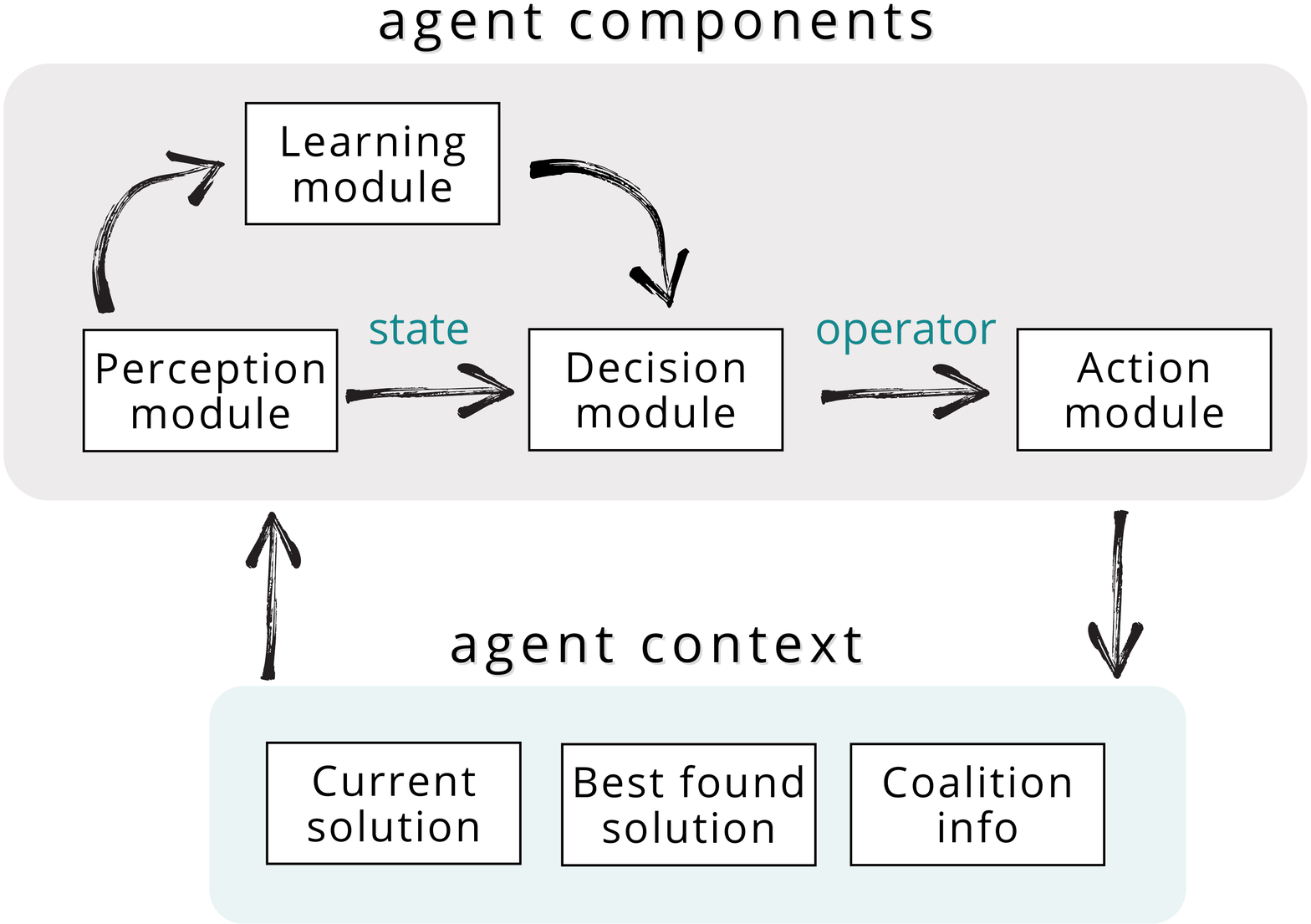}
    \caption{CBM agent structure.} 
    \label{fig:CBM_agent}
\end{figure}

The visual representation of the CBM agents is shown in Figure \ref{fig:CBM_agent}. During the search process, each agent maintains three solutions, similarly to particle swarm optimization \cite{Kennedy95}: a current solution, the best solution found by the agent, and the best solution found by the entire coalition. An agent uses several operators that are applied to the current solution. The operators can be intensifiers or diversifiers. Intensifier operators concern improvement processes such as local search, and diversifier operators correspond to generation, mutation, or crossover procedures.

The choice of operators to apply is not completely stochastic as in GA. Instead, it is determined by a decision process that uses perceived state and past experience to select the most appropriate operators and coordinate intensification and diversification procedures. The selection of operators is based on heuristic rules. The search behavior of an agent is adapted during the optimization process through an individual reinforcement learning mechanism and mimetic learning. These mechanisms modify the rules of the decision process based on the experience results of previous explorations. Although all agents in the coalition use the same set of operators, the learning mechanisms may ultimately lead to different strategies.

Agents cooperate in two ways. First, an agent can inform the rest of the coalition about the newly found best coalition solution. Second, agents share their internal decision rules to enable mimetic behavior. This fosters search behavior in which desirable solutions are often found.

In \cite{Meignan2009}, the authors proposed the CBM solution for a case of VRP. In our case, we consider a C-MDVRP and thus need to formulate a suitable set of operators. Moreover, we modified the basic CBM algorithm to keep more than one current solution so that a population of solutions is preserved, similar to GA methods. In the rest of the paper, we refer to the proposed algorithm as \emph{CBM-pop}. Details on the implementation of the algorithm follow in this section.

\subsection{Distributed Metaheuristic for C-MDVRP}

The first feature to consider in the design of the algorithm is the representation of the solution. Since this metaheuristic is based on a set of operators commonly used in genetic algorithms, we use the same encoding of the solutions in terms of the chromosome. Inspired by an evolutionary process, each chromosome contains genetic material that defines a solution (genotype). In the case of C-MDVRP, this refers to the assignment of actions to different robots and their order within a sequence of tasks in the schedule. Each chromosome is associated with a phenotype that evaluates the genetic material and, in our case, generates schedules for task sequences based on the temporal properties of the tasks.

\begin{figure}[htb]
    \centering
    \includegraphics[clip, trim= 0.1cm -0.5cm 0cm 1.2cm, width=\linewidth]{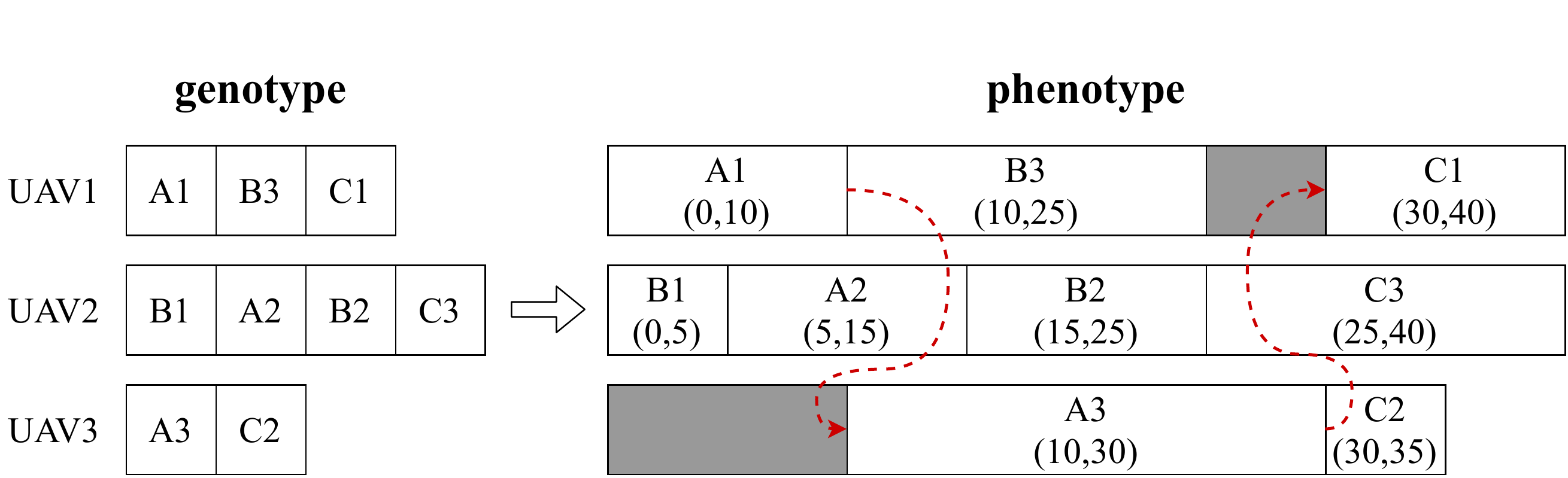}
    \caption{Solution representation -- chromosome genotype and phenotype. The tasks presented here are arbitrarily named generic tasks ($(A,B,C)\times(1,2,3)$). The idle times introduced in the schedules are a consequence of precedence constraints $prec(A1,A3)$ and $prec(A3,C1)$, since task $A3$ cannot start before task $A1$ finishes, and task $C1$ cannot start before the end of $A3$.}
    \label{fig:Chromosome}
\end{figure}

An example of a chromosome and its genotype and phenotype is shown in Figure \ref{fig:Chromosome}. On the left is shown the genetic material of a solution containing specified task groupings of robots (robot1, robot2, robot3) and ordering. The genotype representation is maintained respecting intra-schedule precedence constraints. On the right is an example of a phenotype generated from the specified genotype. The schedule is formed by introducing time elements into the ordered tasks (task durations, task setup times). If necessary, minimal idle times are inserted to ensure consistency with the defined precedence constraints. The phenotype represents the so-called semi-active schedule, where no left shift is possible in the Gantt graph. For any given sequence of robot operations, there is only one semi-active schedule \cite{Sprecher1995}. One advantage of storing solutions in this way is faster exploration of the solution space, since all operators perform on a simpler genotype representation of the solution. The evaluation procedure renders the phenotype and evaluates the solutions found.

The next point to consider is the evaluation of the solution. The usual approach to solution evaluation is to use a fitness function that maps each chromosome in a population to a value of the utility function in $\mathbb{R}$. This usually works best when the search is limited to a single optimization objective. For multi-objective optimization problems, it is best to employ a ranking procedure because these objectives often interact in complex, nonlinear ways. In this work, we use a Pareto ranking procedure \cite{Kalyanmoy2001} that assigns ranks to all solutions based on the non-dominance property (i.e., a solution with a lower rank is clearly superior to solutions with a higher rank concerning all objectives). Therefore, solutions are stratified into multiple ranks based on their ability to meet the optimization objectives.

To evaluate solutions in a population $P$, we apply the double-rank strategy, which takes into account both the density information and the distribution of the solution in the rank. In the first step, an individual $i \in P$ is assigned a dummy rank value $R'(i)$ representing the number of solutions that dominate it in the current population $P$:
\begin{equation}
    R'(i) = |{j, j \in P, i \prec j}|, \forall i \in P,
\end{equation}
where the symbol $\prec$ corresponds to the Pareto dominance relation, i.e., $i \prec j$ if the solution $j$ performs better than $i$ given all optimization criteria. The final rank of solution $R(i)$ is then defined as the sum of its own dummy rank value and that of its dominators:
\begin{equation}
    R(i) = R'(i) + \sum_{j \in P, i \prec j}{R'(j)}, \forall i \in P.
\end{equation}

The second part of the fitness function is the density function, which determines how similar the solution is to the other individuals in the population. Here we use a fairly simple solution where the density of an individual is inversely proportional to the distance to the nearest solution in the population and is calculated as follows:
\begin{equation}
    dens(i) = \frac{1}{min(d(i,j), \forall j \in P) + 2}, \forall i \in P,
\end{equation}
where $d(i,j)$ represents the Euclidean distance between two individuals in the criteria space. Finally, the fitness of the solution is obtained as:
\begin{equation}
\label{eq:Pareto_fitness}
    fitness(i) = \frac{1}{R(i) + dens(i) + 1}, \forall i \in P.
\end{equation}

Next, we briefly discuss the operators that form the core of the algorithm. We distinguish between generation, diversification (crossover and mutation operators), and intensification operators (local search algorithms). In our application, the generation operator is not used as a diversifier because it is applied during initial population creation. In the proposed solution, we use a single generation operator, a greedy insertion method that randomly takes an unassigned task and inserts it into existing routes at minimal cost, taking into account capacity constraints. Other operators are listed and described in Table \ref{tbl:genetic_op}. Diversification operators are first introduced in \cite{Pereira2009}, and we implemented them for our specific problem. In the crossover procedure, we distinguish two cases of Best-Cost Route Crossover (BCRC), depending on the choice of parent chromosomes. One of the parents is always the current solution of the agent and the second parent is either the best solution found within the whole coalition or selected from the population. Similarly, we adapted the local search algorithms developed in \cite{Meignan2009} for a VRP problem class.

\begin{table}[h!]
    \centering
    \caption{Genetic operators used in our proposed solution.}
    \label{tbl:genetic_op}
    \scalebox{0.84}{
    \begin{tabular}{p{0.405\linewidth}p{0.7\linewidth}}
        \hline
        \multicolumn{2}{l}{\textbf{Diversifiers} \cite{Pereira2009}} \\
        \hline
         & \textbf{Crossover} \\
         Best-Cost Route Crossover (BCRC) & For two parent chromosomes, select a route to be removed for each. The removed nodes are inserted into the other parent solution at the best insertion cost. \\
         \rule{0pt}{3ex} & \textbf{Mutation} \\
         intra depot reversal & Two cutpoints in the chromosome associated with the robot initial position are selected and the genetic material between these two cutpoints is reversed. \\
         intra depot swapping & This simple mutation operator selects two random routes from the same initial position and exchanges a randomly selected action from one route to another. \\
         inter depot swapping & Mutation of swapping nodes in the routes of different initial positions. Candidates for this mutation are nodes that are in similar proximity to more than one initial position. \\
         single action rerouting & Re-routing involves randomly selecting one action, and removing it from the existing route. The action is then inserted at the best feasible insertion point within the entire chromosome. \\
        \hline
        \multicolumn{2}{l}{\textbf{Intensifiers} \cite{Meignan2009}} \\
        \hline
         two swap & Swapping of borderline actions from two initial positions to improve solution fitness. \\
         one move & Removal of a node from the solution and insertion at the point that maximizes solution fitness. \\
         \hline
    \end{tabular}  
    }
\end{table}

The behavior of a single CBM-pop agent is described in Algorithm \ref{alg:pop_cbm}. Before starting the algorithm, the agents exchange their specific problem parameters -- task durations $\sigma_{ir}$, setup times $t_{ijr}$, setup costs $c_{ijr}$, energy demands $q_{ir}$, and energy capacities $Q_r$. During the runtime of the algorithm, the best solutions found and the weight matrices are exchanged among the agents, as noted in Algorithm \ref{alg:pop_cbm}. The procedure itself consists of Diversification-Intensification cycles (D-I cycles), where a diversification operator is first applied to the solution to guide the search out of the local optimum. After this perturbation, a series of local search procedures are applied to the solution to arrive at a new (local) optimum. The process is repeated until a termination criterion is reached. Further details on the definition of states, experience updates and learning mechanisms can originally be found in \cite{Meignan2009}.

\begin{algorithm}[h!]
\SetAlgoLined
 \SetKwInOut{Input}{input}
 \SetKwInOut{Output}{output}
 \Input{$pop\_size$ -- number of solutions in population}
 \Input{$\bm{\eta}$ -- reinforcement learning factors}
 \Input{$\rho$ -- mimetism rate}
 \Input{$n\_cycles$ -- number of cycles before changing exploration origin}
 \Input{$\epsilon$ -- minimal solution improvement}
 \Output{$c_{best\ coalition}$ -- best found solution}
 \tcc{initialization}
 $P \gets$ generate\_population($pop\_size$)\\
 evaluate\_population($P$)\\
 $c_{current} \gets$ select\_solution($P$)\\
 $W \gets$ init\_weight\_matrix()\\
 $H \gets$ init\_experience\_memory()\\
 \While{stopping criterion is not reached}{
    \tcc{calculate current state}
    $s \gets$ perceive\_state($H$)\\
    \If{no change in best solution > $\epsilon$ for $n\_cycles$ cycles}{
        evaluate\_population($P$)\\
        $c_{current} \gets$ select\_solution($P$)
    }
    $o$ $\gets$ choose\_operator($W$, $s$)\\
    $c_{new} \gets$ apply\_op($o$, $c_{current}$, $P$, [$c_{best\ coalition}$])\\
    \tcc{update experience history}
    $gain \gets f(c_{current}) - f(c_{new})$\\
    update\_experience($H$, $s$, $o$, $gain$)
    \tcc{update solutions}
    \If{$c_{best\ coalition}$ improved}{
        broadcast\_solution($c_{new}$)
    }
    \tcc{learning mechanisms}
    \If{end of D-I cycle}{
        \uIf{$c_{best\ coalition}$ improved in the cycle}{
            $W$ $\gets$ individual\_learning($W$, $H$, $\bm{\eta}$)
        }
        \ElseIf{$c_{best\ ag}$ improved in the cycle}{
            $W$ $\gets$ individual\_learning($W$, $H$, $\bm{\eta}$)\\
            broadcast\_weight\_matrix($W$)
        }
    }
    \If{weight matrix received from a neighbor}{
    $W$ $\gets$ mimetism\_learning($W$, $W_{received}$, $\rho$)
    }
 }
 \caption{Population-based CBM algorithm (CBM-pop).}
 \label{alg:pop_cbm}
\end{algorithm}

As mentioned, another novelty introduced here is that an agent stores a population of solutions instead of just one solution. The idea is to diversify the search further and allow for broader exploration. The role of the population is a dual one. First, after each $n\_cycles$ cycles without improvement over the best solution found, a new starting solution is randomly selected from the population. Second, solutions from the population participate as a second parent in the crossover operator, thereby introducing novelty from the genetic pool.
\section{Evaluation and Discussion}
\label{sec:results}

In this section, we discuss the performance results of the proposed algorithm on different problem instances compared to the state of the art and optimal solutions. The main properties we are interested in are optimality, computational speed, and scalability. First, we analyze the proposed population-based CBM compared to the original single-solution version of the CBM algorithm. Next, we examine the performance of the proposed CBM-pop on an established set of benchmark examples for MDVRP. We then test the algorithm on a large scale on randomly generated task planning problems with precedence constraints and transitional dependencies. We compare the results with optimal solutions obtained using the Gurobi solver and a state-of-the-art auction-based planner. Finally, we demonstrate the performance of the algorithm in a realistic planning scenario for routing mobile robots in a greenhouse setting.

The proposed solution is implemented in the Robot Operating System (ROS) \cite{ros} environment to enable vehicle-in-the-loop development of planning algorithms. This facilitates the rapid transition from the simulated environment to a real-world experiment. ROS also provides the underlying communication infrastructure through its message and service protocols. All simulations were run on Intel(R) Core(TM) i7-7700 CPU @ 3.60GHz x 8, 32 GB RAM running Ubuntu 18.04 LTS operating system.

\subsection{Performance of the population-based CBM}

\begin{figure}[t!]
    \centering
    \begin{subfigure}{.8\linewidth}
        \centering
      \includegraphics[clip, trim=2cm 0.5cm 4.2cm 2.5cm, width=\linewidth]{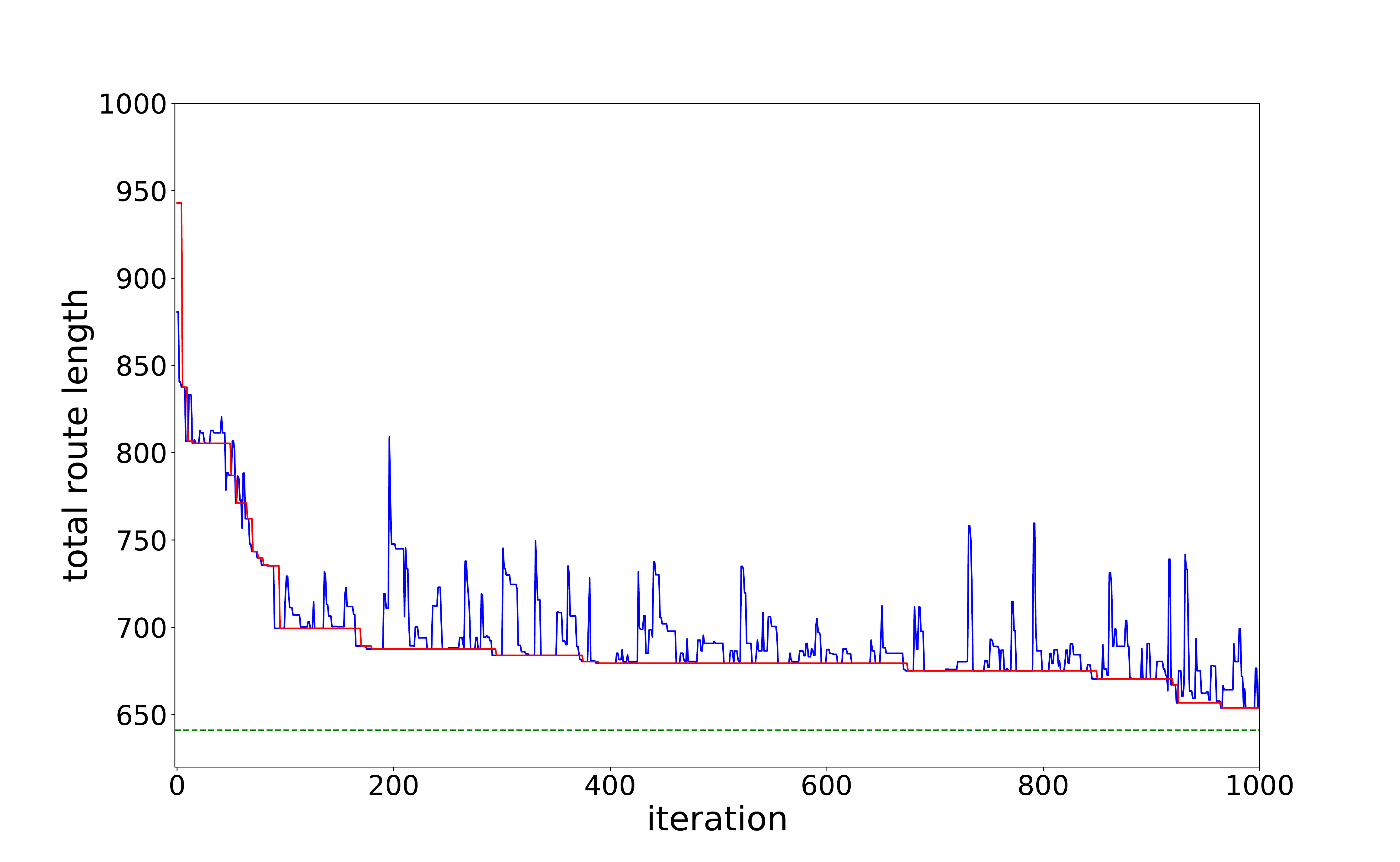}
      \captionsetup{justification=centering}
      \caption{original CBM \cite{Meignan2009}}
      \label{fig:CBM_variants_original}
    \end{subfigure}
    \begin{subfigure}{.8\linewidth}
      \centering
      \includegraphics[clip, trim=2cm 0.5cm 4.2cm 2.5cm, width=\linewidth]{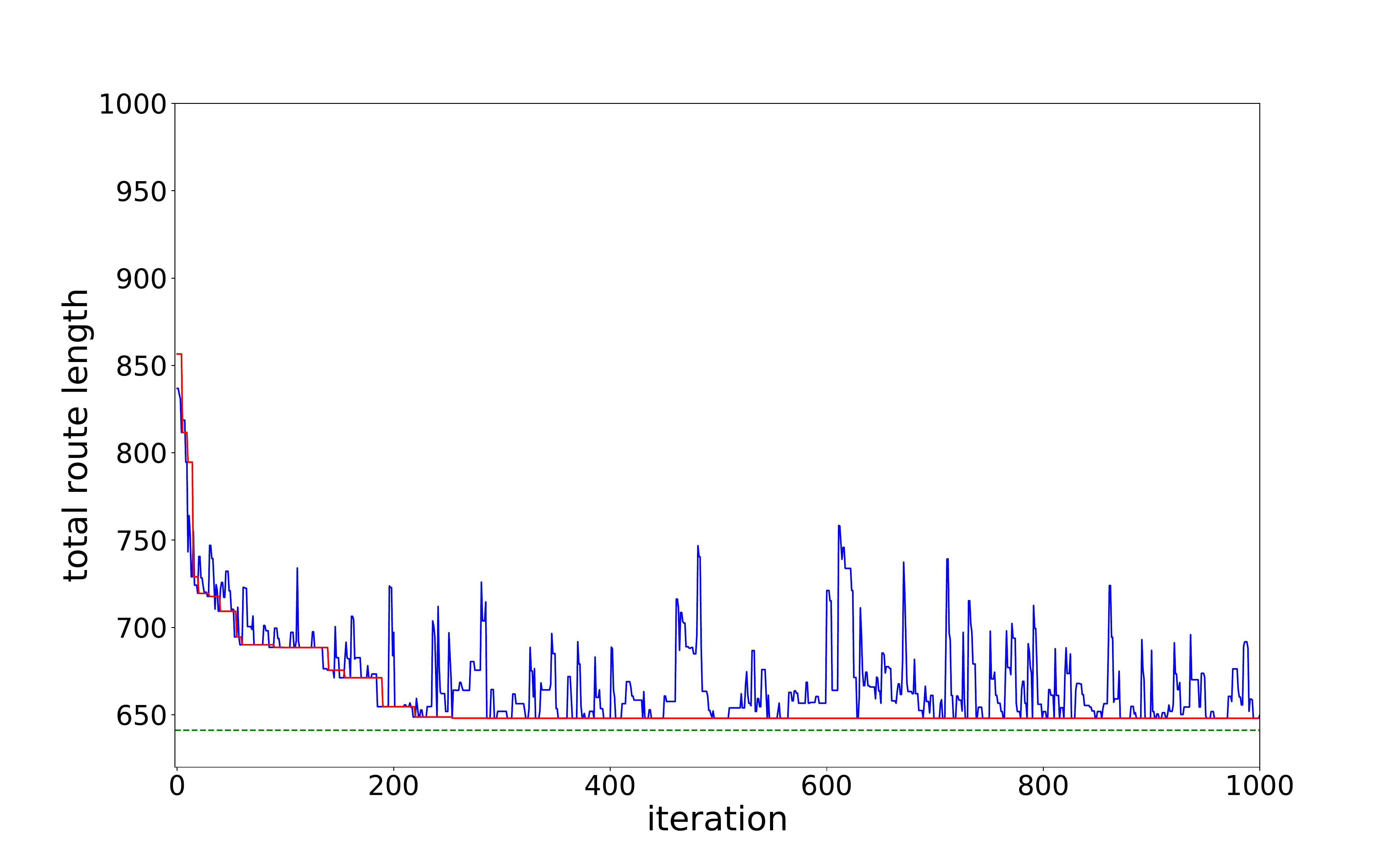}
      \captionsetup{justification=centering}
      \caption{population-based CBM}
      \label{fig:CBM_variants_pop}
    \end{subfigure}
    \caption{Illustration of typical trajectories of the current (blue) and the best found solution (red) for the two variants of the CBM algorithm. We analyze the convergence speed of single-solution CBM and the proposed CBM-pop. The best known solution is shown in the graph with a dashed green line.}
    \label{fig:CBM_variants}
\end{figure}

First, we evaluate the performance of the proposed CBM-pop compared to the single-solution variant of CBM. In Figure \ref{fig:CBM_variants} is the visual representation of the current (blue) and best solution (red) trajectories for the two variants of the CBM algorithm on a benchmark problem for MDVRP. The two plots illustrate the contrast of the two approaches in the early search of the solution space. The CBM-pop manages to quickly jump through various solution configurations and explore different local optima, leading to faster convergence towards the optimal region. In Figure \ref{fig:CBM_variants_pop}, we can observe a rapid convergence (within the first $200$ iterations of the algorithm) of the found solutions towards the dashed green line, which is the best known solution for the given problem. In contrast, the solution trajectories in Figure \ref{fig:CBM_variants_original} show a noticeably slower progress towards better solutions. This difference is the direct result of the genetic diversity of the obtained solution pool of CBM-pop and is consistent with the intended algorithm design.

\subsection{Performance on benchmark examples of MDVRP}

\begin{table*}[htb]
    \caption{Performance comparison between CBM-pop and two state-of-the-art optimization algorithms TSH \cite{HesamSadati2020} and CoEV \cite{deOliveira2016} on Cordeau benchmark examples \cite{CordeauDataset} for MDVRP. In the table, $m$ and $n$ represent the number of vehicles and customers, respectively, for each test example. The best solutions found for all algorithms are given, as well as the optimality gap ($\%$) with respect to the best known solutions (BKS) for the benchmark. The average optimality gap is given for CBM-pop based on 50 runs of the algorithm. The total CPU times are given for algorithms in the defined simulation setup. We estimate the computation time per vehicle $\tau (s)$ when CBM-pop is run fully distributively on each of the vehicles in the system. The best solutions are highlighted in bold.}
    \label{tbl:benchmark_res}
    \scalebox{0.85}{
    \begin{tabular*}{1.14\textwidth}{*{3}{C{0.03\textwidth}}C{0.07\textwidth}*{3}{C{0.055\textwidth}}p{0.5px}*{3}{C{0.055\textwidth}}p{0.5px}*{5}{C{0.055\textwidth}}}
        \hline
        \multirow{3}{*}{test} & \multirow{3}{*}{\centering m} & \multirow{3}{*}{\centering n} & \multirow{3}{*}{BKS} & \multicolumn{3}{c}{TSH\cite{HesamSadati2020}} & & \multicolumn{3}{c}{CoEV\cite{deOliveira2016}} & &
        \multicolumn{5}{c}{CBM-pop} \\
        \cline{5-7} \cline{9-11} \cline{13-17} 
        & & & & best & gap($\%$) & CPU(s) & & best & gap($\%$) & CPU(s) & & best & gap($\%$) & avg. gap($\%$) & CPU(s) & $\tau(s)$\\
        \hline
        p01 & 16 & 50 & \textbf{576.87} & \textbf{576.87} & 0 & 9.4 & & \textbf{576.87} & 0 & 1.0 & & \textbf{576.87} & 0 & 2 & 18.1 & 4.5 \\ 
        p02 & 8 & 50 & \textbf{473.53} & \textbf{473.53} & 0 & 20.9 & & 473.87 & 0 & 0.5 & & \textbf{473.53} & 0 & 1 & 12.3 & 6.2 \\ 
        p03 & 15 & 75 & \textbf{641.19} & \textbf{641.19} & 0 & 141.9 & & \textbf{641.19} & 0 & 2.5 & & \textbf{641.19} & 0 & 1 & 26.1 & 7.0 \\ 
        p05 & 10 & 100 & \textbf{750.03} & 758.87 & 1.16 & 159.1 & & 750.11 & 0 & 26.6 & & 752.05 & 0.27 & 2 & 43.1 & 17.2 \\ 
        p06 & 18 & 100 & \textbf{876.50} & 881.76 & 0.60 & 194.8 & & \textbf{876.50} & 0 & 77.3 & & 893.59 & 1.91 & 3 & 44.0 & 9.8 \\
        p09 & 36 & 249 & 3900.22 & 3971.59 & 1.80 & 606.0 & & \textbf{3895.70} & \textbf{-0.12} & 513.30 & & 4049.60 & 3.69 & 9 & 97.7 & 10.9 \\ 
        p10 & 32 & 249 & \textbf{3663.02} & 3779.10 & 3.07 & 703.7 & & 3666.35 & 0 & 719.90 & & 3792.10 & 3.40 & 8 & 117.3 & 14.7 \\ 
        p11 & 30 & 249 & \textbf{3554.18} & 3652.01 & 2.68 & 660.3 & & 3569.68 & 0.43 & 396.20 & & 3713.39 & 4.29 & 8 & 122.1 & 16.3 \\ 
        p12 & 10 & 80 & \textbf{1318.95} & \textbf{1318.95} & 0 & 13.8 & & \textbf{1318.95} & 0 & 0.9 & & \textbf{1318.95} & 0 & 1 & 24.1 & 9.7 \\ 
        p13 & 10 & 80 & \textbf{1318.95} & \textbf{1318.95} & 0 & 6.7 & & \textbf{1318.95} & 0 & 0.0 & & \textbf{1318.95} & 0 & 0 & 24.6 & 9.9 \\ 
        p15 & 20 & 160 & \textbf{2505.42} & 2552.79 & 1.86 & 255.4 & & \textbf{2505.42} & 0 & 432.0 & & 2565.40 & 2.34 & 4 & 57.6 & 11.5 \\ 
        p18 & 30 & 240 & \textbf{3702.85} & 3802.29 & 2.62 & 302.9 & & 3771.35 & 1.82 & 429.1 & & 3814.23 & 2.92 & 6 & 159.6 & 21.3 \\ 
        p21 & 45 & 360 & \textbf{5474.84} & 5617.53 & 2.54 & 1703.0 & & 5608.26 & 2.38 & 554.9 & & 5731.88 & 4.48 & 7 & 267.3 & 23.8 \\ 
        pr01 & 4 & 48 & \textbf{861.32} & \textbf{861.32} & 0 & 2.1 & & \textbf{861.32}  & 0 & 0.0 & & \textbf{861.32} & 0 & 0 & 4.8 & 4.8 \\ 
        pr09 & 18 & 216 & 2153.10 & 2177.20 & 1.11 & 613.5 & & \textbf{2150.52} & \textbf{-0.12} & 1107.15 & & 2187.84 & 1.59 & 5 & 75.5 & 16.8 \\ 
        \hline
    \end{tabular*}
    }
\end{table*}

Given the compelling similarity between the problems of C-MDVRP and task planning, we first analyze the performance of the developed algorithm using well-studied benchmark examples of MDVRP. The benchmark problems we regard \cite{Cordeau97, CordeauDataset} require the assignment of customers to depots and the routing of individual vehicles. The problem solution must serve all customers while minimizing the distance traveled. There is also a maximum route capacity assigned to each vehicle.


We compare the performance of the CBM-pop algorithm with two recent operations research papers that address the same problem on the Cordeau benchmark dataset. The first algorithm selected is the Tabu Search heuristic (TSH) algorithm presented in \cite{HesamSadati2020}. The TSH algorithm was chosen for comparison as a centralized state-of-the-art heuristic algorithm from the field of operations research. The other selected algorithm is a distributed cooperative coevolutionary algorithm (CoEV) \cite{deOliveira2016}. This algorithm runs in a distributed manner, where for each depot a partial solution is constructed and optimized. The TSH algorithm was run on Intel Xeon (R) E5-2690 v4 x 14 processor (2.60 GHz, 32 GB RAM), while the CoEV was run on a computer with two 2.50 GHz Intel Xeon (E5-2640) processors with 12 cores per CPU and 96 GB RAM.

CBM-pop is a distributed algorithm that should typically run on each robot during the mission runtime. For comparison with the TSH and CoEV algorithms, the CPU times in Table \ref{tbl:benchmark_res} are the times required to compute the contribution of all robots in CBM-pop on only one computer. Therefore, CPU times reflect the total computational effort for all algorithms. On the other hand, $\tau (s)$ is the runtime per vehicle when CBM-pop is executed distributively in each vehicle. Thus, the runtimes of CBM-pop are given by $\tau$, while for the other two algorithms they are given by the CPU times. CoEV simulations are originally run fully distributively, and the given CPU times reflect the actual performance of this algorithm.

The full simulation results are shown in Table \ref{tbl:benchmark_res}. The benchmark examples range from $2$ to $6$ for the number of depots, with $2$ to $14$ vehicles per depot, and customer numbers from $48$ to $360$. In Table \ref{tbl:benchmark_res}, the performance of algorithms is directly related to the best known solutions (BKS) for each test example in \cite{CordeauDataset}, and the optimality gap for best simulation runs is also given.

We can conclude that the algorithms CBM-pop and TSH give very similar results in terms of optimality, both well within $5\%$ of the best solutions found for the problem. However, since CBM-pop is distributed, the total CPU effort is shared between all involved robots, resulting in significantly lower runtimes (given by $\tau(s)$). On the other hand, CoEV shows the best performance in terms of optimality of solutions. Although this algorithm is distributed, it shows significantly higher runtimes for examples with more vehicles and customers. The possible reason is that the subproblems become more complex, so the distribution of the problem per depot shows less influence on the performance of the algorithm. In summary, our proposed algorithm CBM-pop produces solutions that are on par in optimality with state-of-the-art operations research approaches to this problem. Due to the distributed nature of CBM-pop, it generates solutions in less runtime, which makes it suitable for online applications for mission planning problems. Another major advantage of CBM-pop is its robustness to agent breakdowns since the computations are performed on multiple nodes.

These tests indicate that CBM-pop provides a very good solution to the given problem and does so in a short computation time. It is important to note that these benchmarks were developed as a very complex test for optimization algorithms and that the time taken to compute the solutions is usually not crucial. For us, the timeliness of the solutions is essential since we are dealing with implementation on a robotic system. With this test, we have shown that our distributed algorithm can handle cases with higher complexity than we expect in robotic systems in a relatively short time. Further simulation results and analysis are available on the paper webpage \cite{PaperWebpage}.


\subsection{Comparative analysis on tasks with cross-schedule precedence constraints}

Despite the considerable complexity of the previous tests, they do not include some important elements of the problem that our approach solves, namely cross-schedule dependencies. Therefore, we have developed a separate set of benchmark examples to evaluate problems with precedence constraints. The problems are solved for a team of two and eight robots. The problem instances include $4$, $8$, $16$, $32$, $64$, $128$, $256$, $512$, and $1024$ tasks, and we generated $50$ randomized examples for each setting. In each example, $20\%$ of the tasks are precedence constrained. Task durations and costs are generated using assumed robot characteristics (speed, energy requirements). Setup times and costs are calculated using the above mentioned agent properties and the Euclidean distance between tasks associated with random positions in 3D space. The full benchmark set and results can be found at the benchmark website \cite{BenchmarkRepo}.

\begin{figure*}[htb]
    \centering
    \captionsetup[subfigure]{justification=centering}
    \begin{minipage}{0.02\linewidth} \hspace{-0.2cm} \vspace{0.5cm} \rotatebox{90}{2 robots}\end{minipage}
    \begin{subfigure}{0.32\textwidth}
        \includegraphics[clip, trim= 2.4cm 0.8cm 4cm 2cm, width=\linewidth]{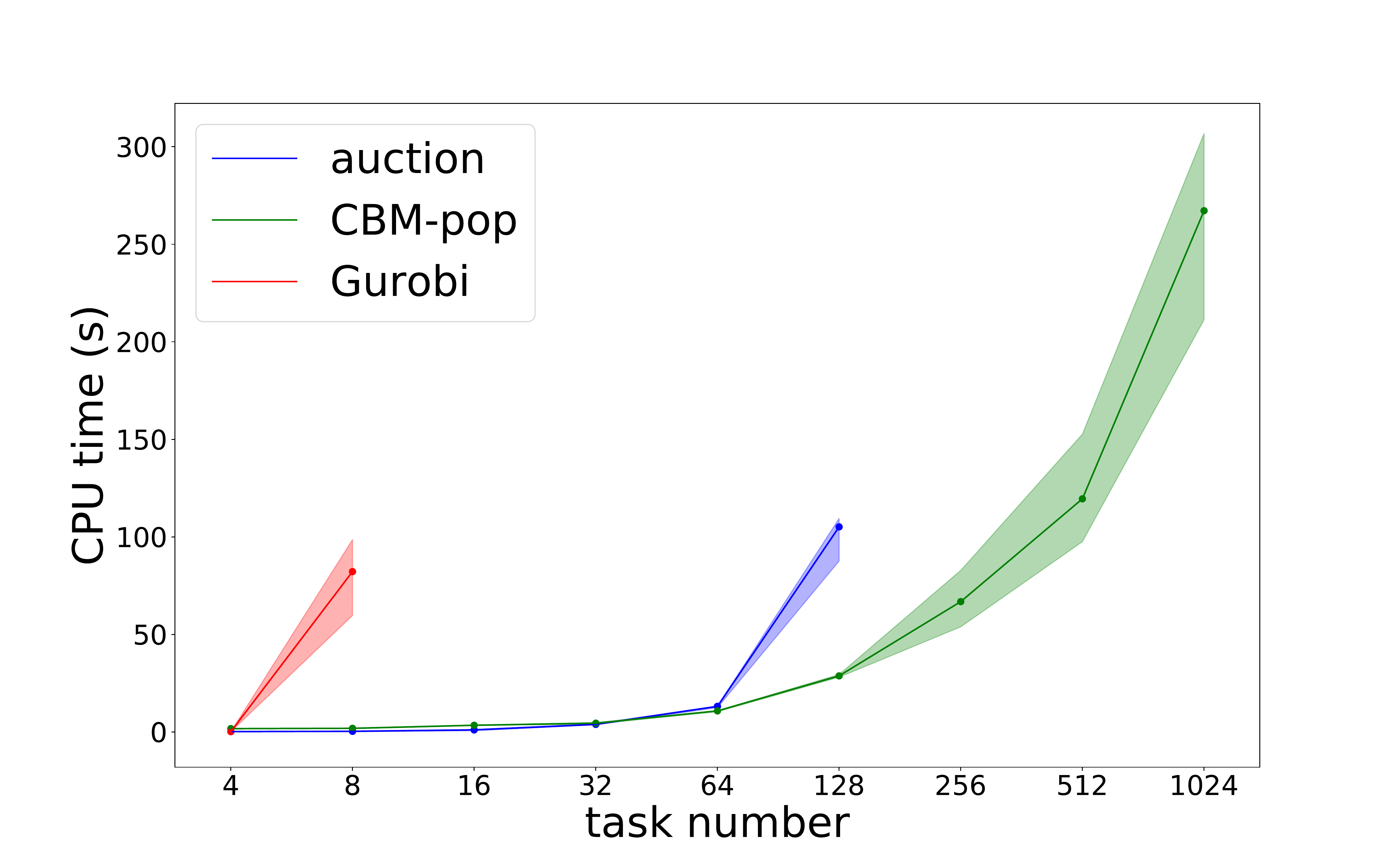}
        \caption{CPU time (s)}
    \end{subfigure}\hfil
    \begin{subfigure}{0.32\textwidth}
      \includegraphics[clip, trim= 1.6cm 0.8cm 4cm 2cm, width=\linewidth]{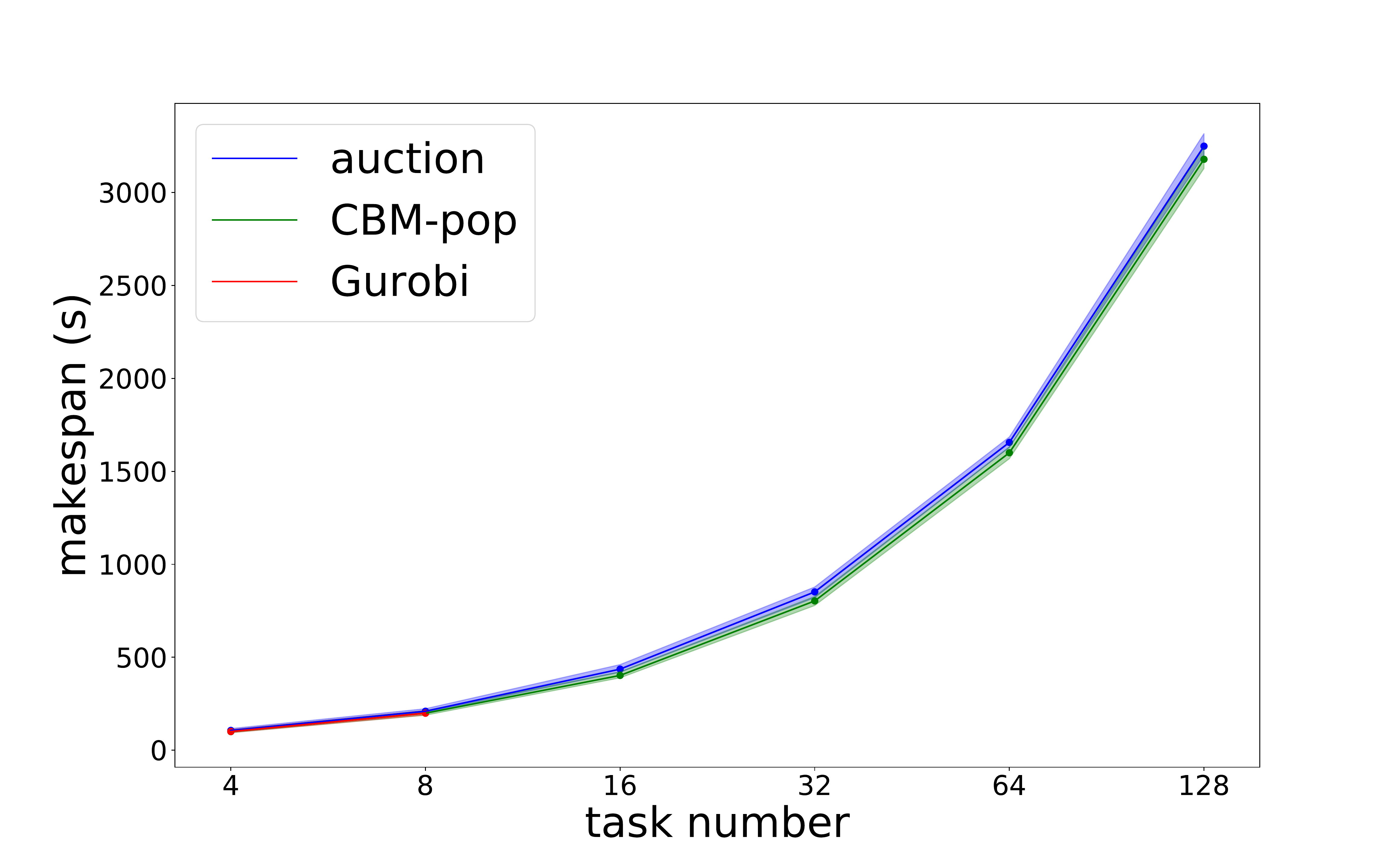}
      \caption{makespan (s)}
    \end{subfigure}\hfil
    \begin{subfigure}{0.32\textwidth}
      \includegraphics[clip, trim= 0.6cm 0.8cm 4cm 2cm, width=\linewidth]{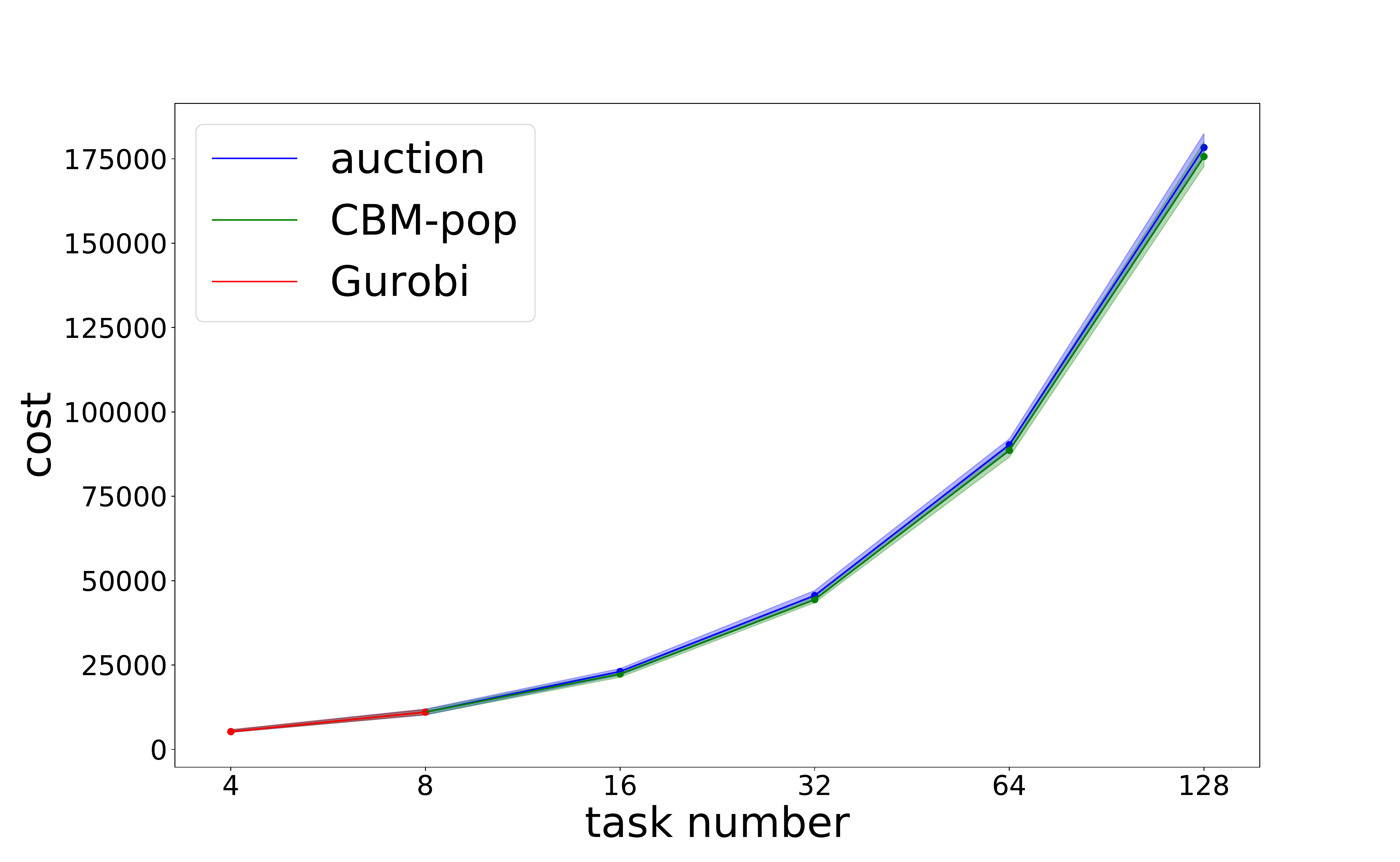}
      \caption{cost}
    \end{subfigure}
    \begin{minipage}{0.02\linewidth} \hspace{-0.2cm} \vspace{0.5cm} \rotatebox{90}{8 robots}\end{minipage}
    \begin{subfigure}{0.32\textwidth}
      \includegraphics[clip, trim= 2.4cm 0.8cm 4cm 2cm, width=\linewidth]{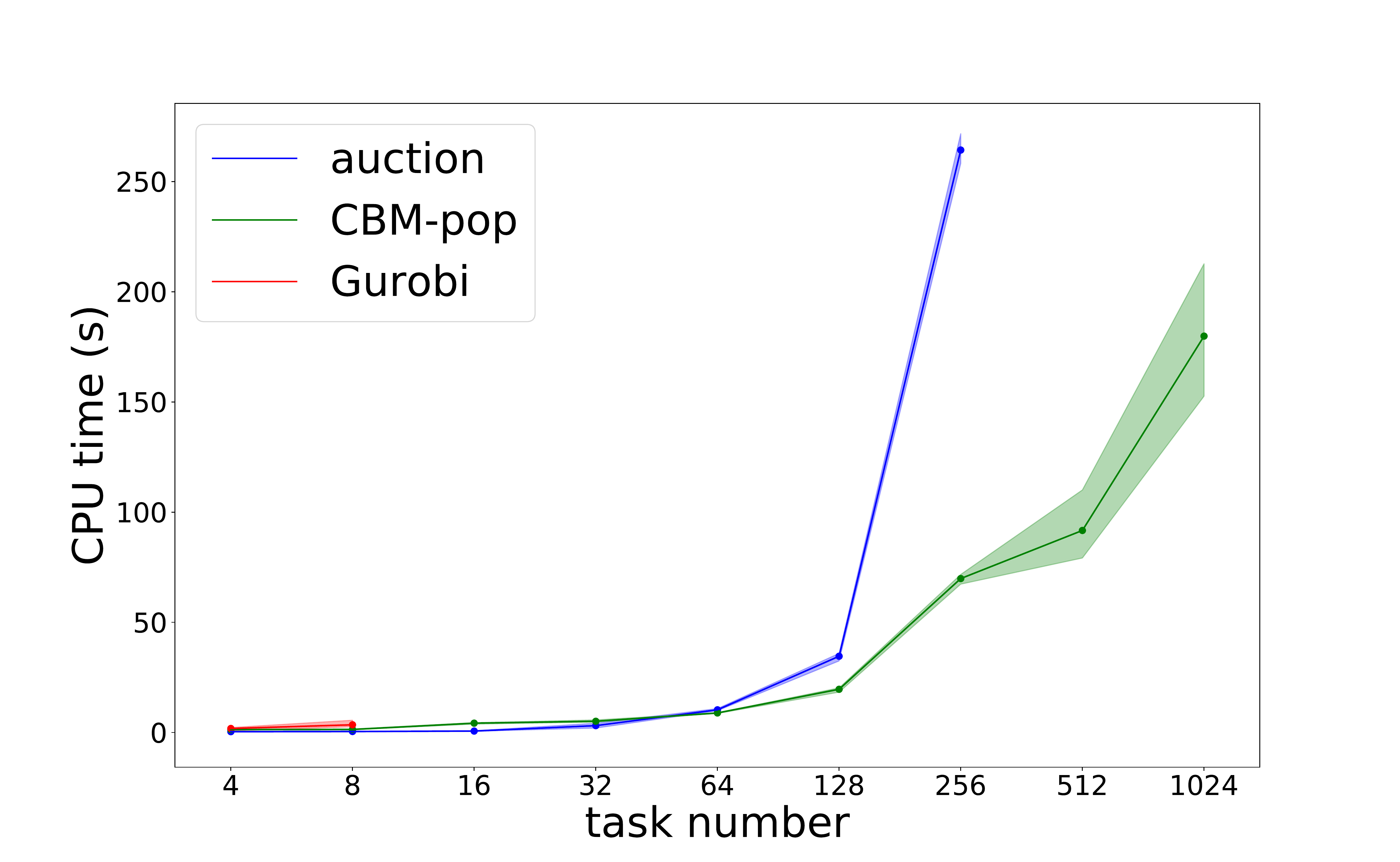}
      \caption{CPU time (s)}
    \end{subfigure}\hfil
    \begin{subfigure}{0.32\textwidth}
      \includegraphics[clip, trim= 1.6cm 0.8cm 4cm 2cm, width=\linewidth]{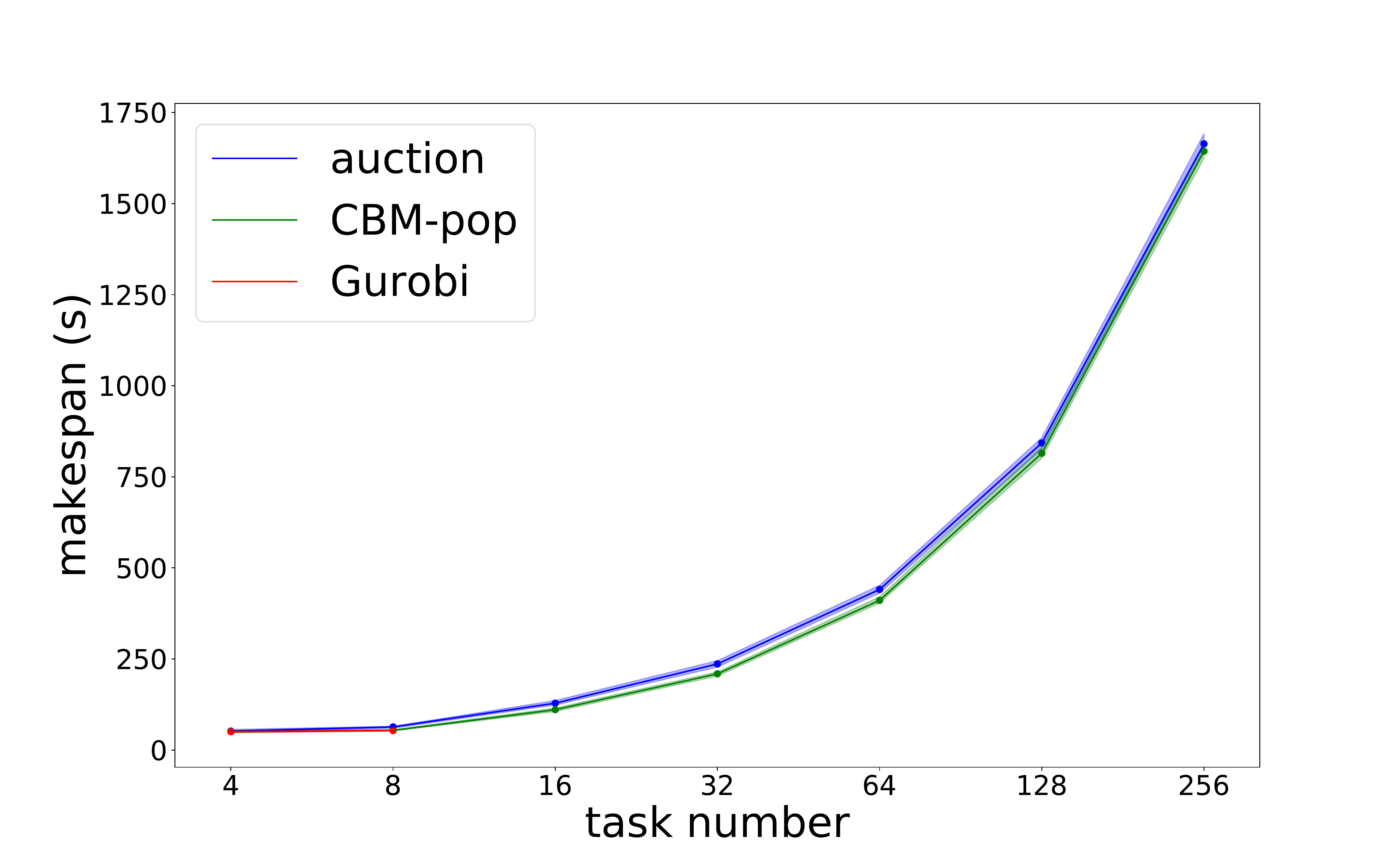}
      \caption{makespan (s)}
    \end{subfigure}\hfil
    \begin{subfigure}{0.32\textwidth}
      \includegraphics[clip, trim= 0.6cm 0.8cm 4cm 2cm, width=\linewidth]{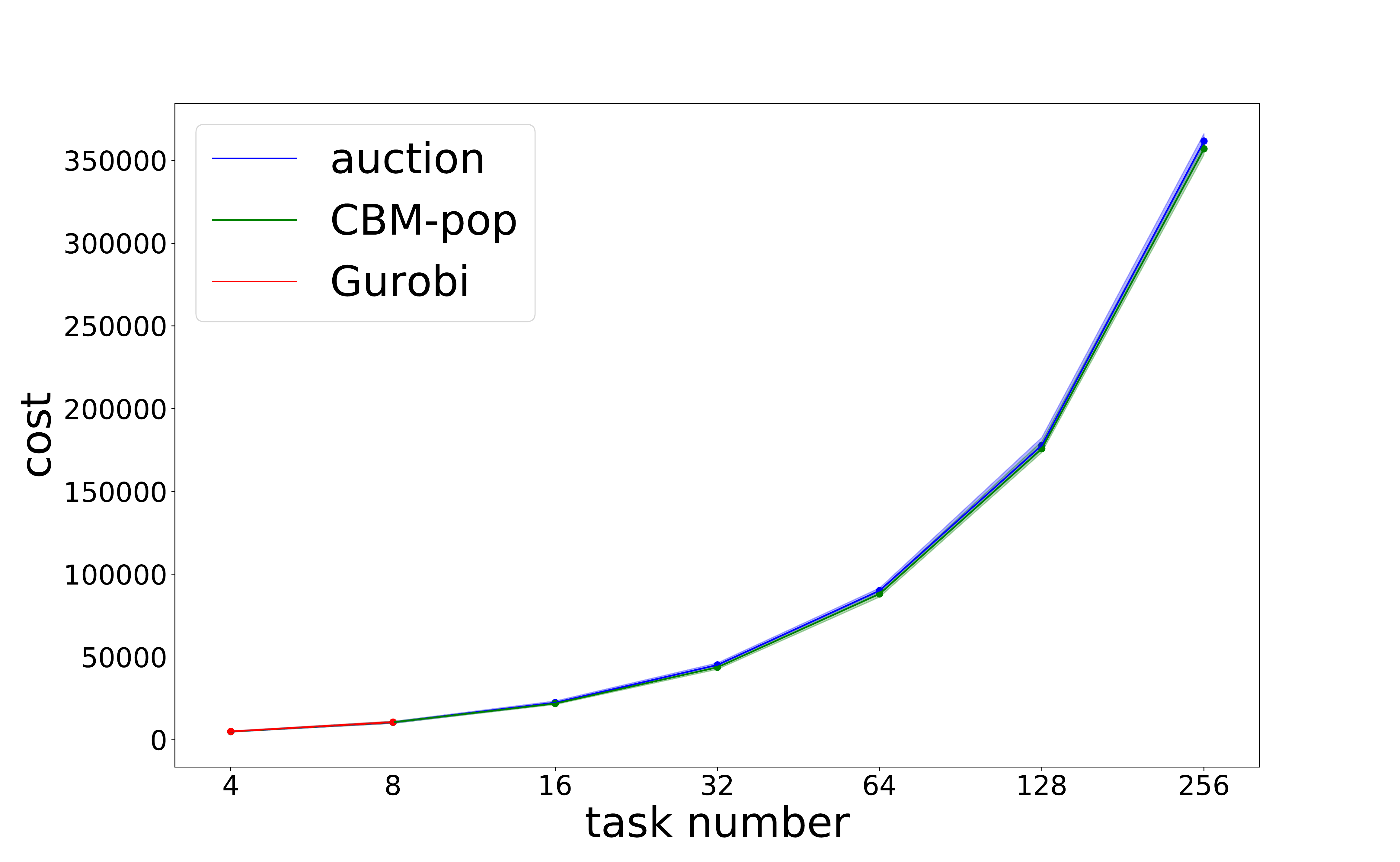}
      \caption{cost}
    \end{subfigure}
    \caption{Comparison of the performance of the proposed distributed metaheuristic algorithm CBM-pop (green), the Gurobi optimal solver (red), and the state-of-the-art auction-based distributed algorithm \cite{Nunes2017} (blue). The top row shows the results for $2$ robots, and the performance for $8$ robots is shown in the bottom row. The plot displays a graph for the mean of each of the observed values and the distribution highlighted by a transparent ray around the graph line. A time limit of $20 min$ was introduced in the simulations, and only the results obtained in this runtime are presented.}
    \label{fig:planning-benchmark-results}
\end{figure*}

In analyzing the results, we compare the proposed distributed metaheuristic CBM-pop with an exact MILP-based solution provided by the Gurobi solver \cite{gurobi} and a state-of-the-art distributed auction-based algorithm presented in \cite{Nunes2017}. The exact solver provides optimal solutions and is used as a basis for solution optimality. However, it cannot compute solutions for larger test examples, and for these we can only observe the similarity of the two heuristic algorithms. The performance results of the algorithm for the defined benchmark set are shown in Figure \ref{fig:planning-benchmark-results}.

Two rows of the figure represent the performance of the algorithms for teams of $2$ and $8$ robots, respectively. In the simulations, we introduced a computation time limit of $20$ minutes. The optimal solver is able to obtain solutions for up to $8$ tasks, the auction-based algorithm, for up to $128$ tasks for the case of $2$ robots, and $256$ tasks for $8$ robots, while the proposed algorithm can handle all examples in the benchmark. For the case of $2$ robots and $256$ tasks, and $8$ robots and $512$ tasks, the auction algorithm takes about $25$ minutes to produce solutions.

The first property to be observed is the algorithm runtime and scalability of the above approaches. From the first column of the grid in Figure \ref{fig:planning-benchmark-results}, it is clear how fast the combinatorial explosion manifests in the auction-based algorithm. It is even clearer for the optimal solver. The Gurobi solver succeeds on problems with up to $8$ tasks for both sets of benchmarks. The auction method is able to solve problems with at most $256$ tasks in the given time. An exponential increase in computation time can be observed. On the other hand, CBM-pop copes very well with an increase in the number of robots and tasks. Also, a larger scatter in the CPU time in CBM-pop is observed for larger task examples. This is due to the stochastic nature of the protocol and the quality-based stopping criterion, which terminates the computation if no improvement in the solution has been achieved for a certain number of steps.

\begin{table*}[h!]
    \centering
    \caption{Summary of average improvements of the proposed algorithm (CBM-pop) compared to the state-of-the-art auction-based algorithm \cite{Nunes2017} on a basis of $50$ randomized examples for each problem setting.}
    \label{tbl:tpl_benchmark_res}
    \begin{tabular*}{0.8\textwidth}{C{0.05\textwidth}C{0.15\textwidth}*{3}{C{0.15\textwidth}}}
        \cline{2-5}
        & \multirow{2}{0.1\textwidth}{\centering task number} & \multicolumn{3}{c}{average improvement ($\%$)} \\
        \cline{3-5}
        & & makespan & cost & CPU time \\
        \cline{2-5}
        \multirow{7}{*}{\rotatebox[origin=c]{90}{2 robots}} & \multicolumn{1}{c}{$4$} & $4.36$ & $0$ & - \\
        & \multicolumn{1}{c}{$8$} & $5.25$ & $0.51$ & - \\
        & \multicolumn{1}{c}{$16$} & $7.45$ & $3.12$ & - \\
        & \multicolumn{1}{c}{$32$} & $5.03$ & $2.36$ & - \\
        & \multicolumn{1}{c}{$64$} & $3.39$ & $1.86$ &  $15.89$\\
        & \multicolumn{1}{c}{$128$} & $2.39$ & $1.57$ & $71.60$ \\
        \cline{2-5}
        \multirow{7}{*}{\rotatebox[origin=c]{90}{8 robots}} & \multicolumn{1}{c}{$4$} & $4.57$ & $0$ & - \\
        & \multicolumn{1}{c}{$8$} & $14.39$ & $0$ & - \\
        & \multicolumn{1}{c}{$16$} & $15.04$ & $2.57$ & - \\
        & \multicolumn{1}{c}{$32$} & $11.48$ & $2.70$ & - \\
        & \multicolumn{1}{c}{$64$} & $7.00$ & $1.93$ & $12.76$ \\
        & \multicolumn{1}{c}{$128$} & $3.43$ & $1.27$ & $42.35$ \\
        & \multicolumn{1}{c}{$256$} & $1.32$ & $1.1$ & $73.11$ \\
        \cline{2-5}
    \end{tabular*}
\end{table*}

In Table \ref{tbl:tpl_benchmark_res}, we can observe the average improvement of the CPU time of the proposed algorithm compared to the auction-based method. We provide information for more extensive problems where the qualities of the proposed method are highlighted. For simpler examples, the auction-based algorithm renders solutions faster (about $2$-$3$ times). CBM-pop computes solutions in about $1.5s$ for the simplest cases ($5s$ for more complex problems), and the auction manages to solve the problem in approximately $0.5s$ and $2s$, respectively.

We also investigate the performance of the algorithms in terms of optimality. As explained above, we model the optimization function in terms of Pareto optimality (as defined in Eq. \ref{eq:Pareto_fitness}) for two criteria, namely the makespan of the schedule and the total cost (Eq. \ref{eq:unified-model-obj-min}). For the limited example set with computed optimal solutions, both the auction and our method follow the optimal solutions very closely (well within $0.5\%$ of the optimum). In Figure \ref{fig:planning-benchmark-results}, the last two columns represent the duration and cost of the solution found. In all examples presented here, the CBM-pop algorithm outperforms the auction for both given criteria. For the case of $2$ robots, the improvements range in makespan from $2\%$ to $7\%$ and in cost up to $3.12\%$. For $8$ robots, the improvements range in makespan from $2$-$16\%$ and in cost up to $2.6\%$. All results are summarized in Table \ref{tbl:tpl_benchmark_res}.

By running these simulations, we have shown that our approach can keep up with the current state of the art in task planning in terms of optimality, while generating solutions in significantly less time, which is essential for all real-world applications.

\subsection{Application to the use-case in agricultural environment}

The use of robots in agriculture is not a new concept, but rather a rapidly growing industry that focuses primarily on large machines used for specific crops and applications. However, the main objective of the SpECULARIA project is to develop a heterogeneous robotic system consisting of three types of agents: an Unmanned Aerial Vehicle (UAV), an Unmanned Ground Vehicle (UGV), and a compliant manipulator with several degrees of freedom. 

The primary role of the UAV in the system is to monitor the system, where it can issue possible maintenance actions. The UGV is equipped with a mechanism that allows it to transport containers of growth units, which are the smallest organizational unit within the farm and consist of a single plant or a variety of plants. The task of the compliant manipulator is to perform delicate manipulations on plants, such as handling flowers and fruits and pruning plants. Each robot has several specific capabilities, but when used together, they can be used in different ways to accomplish multiple objectives.

The mission given to our robotic team is to perform daily maintenance tasks in a robotized greenhouse. The team consists of three unmanned ground vehicles (UGVs) equipped with a mechanism to transport plant containers to and from the workspace and a single robotic manipulator that performs actions on plants. Each UGV can pick up and transport one plant at a time and place them on specific tray holders in the greenhouse. We assume that missions to the system are issued in time frames greater than the time required to complete a single mission, i.e., the team has enough time to complete one mission before the next one arrives.

\begin{figure}[htb]
    \centering
    \includegraphics[width=0.75\linewidth]{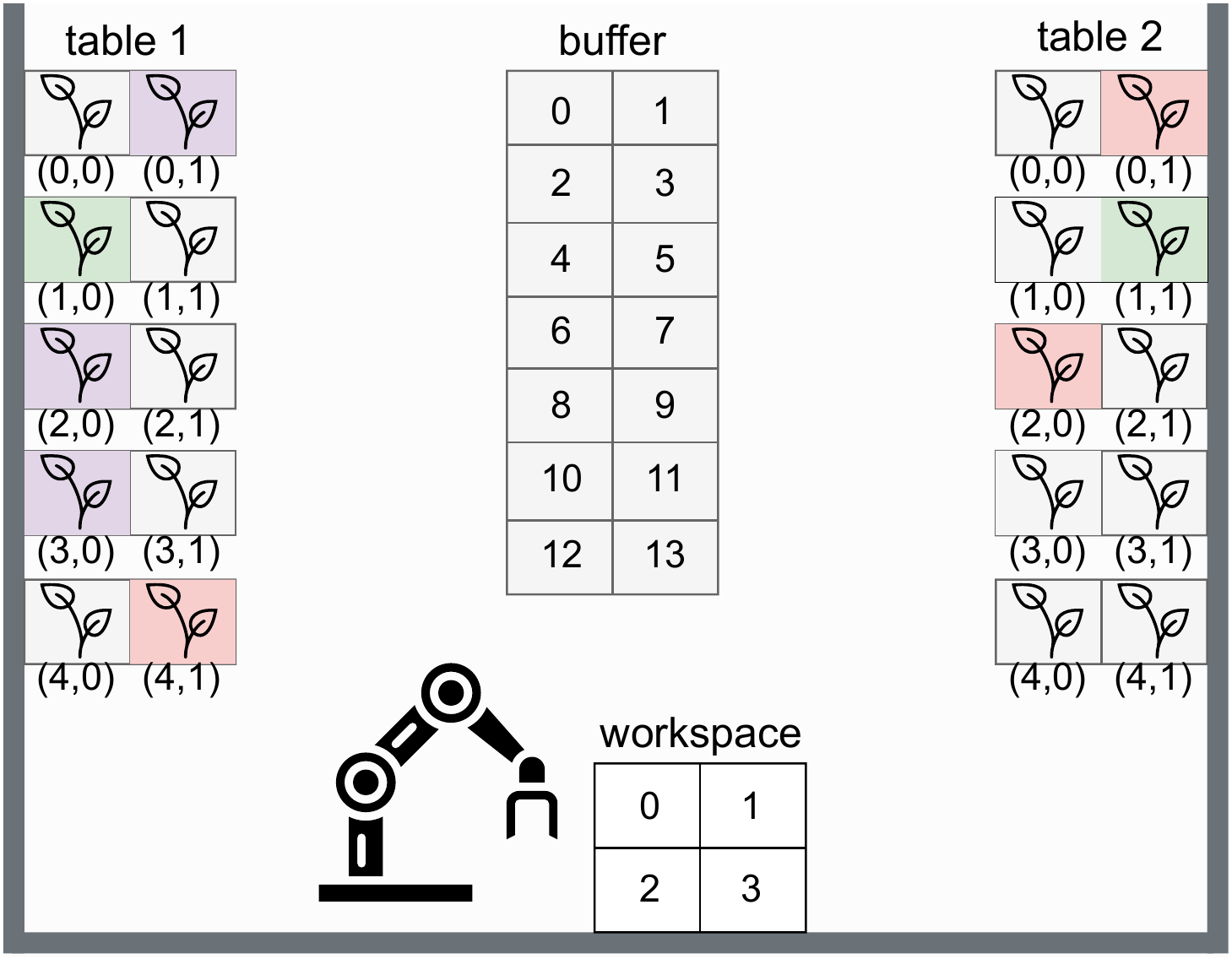}
    \caption{SpECULARIA use case greenhouse layout.}
    \label{fig:greenhouse_layout}
\end{figure}

The layout of a greenhouse in a mission we analyze in this paper is shown in Figure \ref{fig:greenhouse_layout}. The greenhouse consists of two tables with plants placed along the walls of the greenhouse. Each table is organized into five rows and two columns. The tables are enumerated, as are the specific positions within the table. The convention for addressing the tray holders within the table is \emph{(row, column)} and the indices begin with $0$. The full address of each plant is defined by the triplet \emph{(table, row, column)}. Plants within the table can only be accessed by row, starting with the positions at the aisle. Thus, plants located by the wall of the structure can only be accessed by removing previous plants in the same row of the table. For example, in Figure \ref{fig:greenhouse_layout}, the plant at position $(1,0)$ in table $1$ can only be accessed after removing the plant at position $(1,1)$ from the table. A similar precedence relationship applies to table $2$, except that plants are accessed from left to right.

In addition to the two tables, there is a buffer table structure in the middle of the greenhouse. The structure of a buffer is very similar to the structure of the tables, but the plants can be reached from either side, so there is no precedence relationship between the plant access tasks. The buffer is used to store plants that need to be put aside before the required plants are transported to the processing station (work station of the robot manipulator). Plants that are finished with the maintenance task are also put back into the buffer.

Finally, at the bottom of the greenhouse structure is a workspace table with four plant tray holders. The idea of the four positions is to allow for batch processing of the plants, which is especially advantageous for simpler tasks such as watering or spraying the plants.

Inputs to the planning procedure include the greenhouse layout, the groups of plants to be tended that day, and the procedures to be performed. In the given example from Figure \ref{fig:greenhouse_layout}, the specific groups are $A = \{(1,0,1),\allowbreak (1,2,0),\allowbreak (1,3,0)\}$, marked in purple in the figure, $B = \{(1,1,0),\allowbreak (2,1,1)\}$ marked in green, and $C = \{(1,4,1),\allowbreak (2,0,1),\allowbreak (2,2,0)\}$ marked in red. This means that to execute operation $A$, all three defined plants must be present in the workspace table. The same is true for the other two tasks.

In problem modeling, we distinguish between two actions that the UGV can perform on plants, namely transporting plants to the buffer and moving plants to the workspace. Based on these two actions and the defined inputs, we generate a set of actions to be planned for. For plants that are not scheduled for care on a given day and that interfere with the plants to be processed, the action of moving them to the buffer is generated. For example, we can identify the task \emph{to\_buffer(1,1,1)}. For plants that are scheduled for care, we define two precedence-constrained actions, moving them to the workspace and placing them in the buffer when processing is complete. For example, we define the tasks \emph{to\_workspace(1,1,0)} and \emph{to\_buffer(1,1,0)}, with a precedence constraint in between. Additionally, there is a precedence for the tasks of accessing two adjacent plants, in this example the tasks \emph{to\_buffer(1,1,1)} and \emph{to\_workspace(1,1,0)}.

On the side of the manipulator that tends the plants, we define three different actions. This was necessary to ensure the desired system behavior while keeping the problem within the scope of the defined modeling. For the maintenance task \emph{A} example, the defined tasks are \emph{A\_ready}, \emph{A\_perform}, and \emph{A\_setup}, all of which take precedence in the defined order. Task \emph{A\_ready} signals that the workspace is empty and ready to receive the next batch of plants. This task precedes all tasks \emph{to\_workspace} for the given procedure. After all plants are placed on the workspace, the task to perform the procedure (\emph{A\_perform}) begins. This relationship is also modeled by precedence constraints. Next, after the procedure is completed, the tasks of transporting the plants from the workspace to the buffer are activated. After all plants are removed from the workspace, the task \emph{A\_setup} is executed. This task represents the tool change of a robot arm. When it is finished, new plants can be brought into the work area and the whole process starts again.

\begin{figure}[htb]
    \centering
    \includegraphics[clip,trim=5cm 4cm 3.5cm 2cm,width=0.9\linewidth]{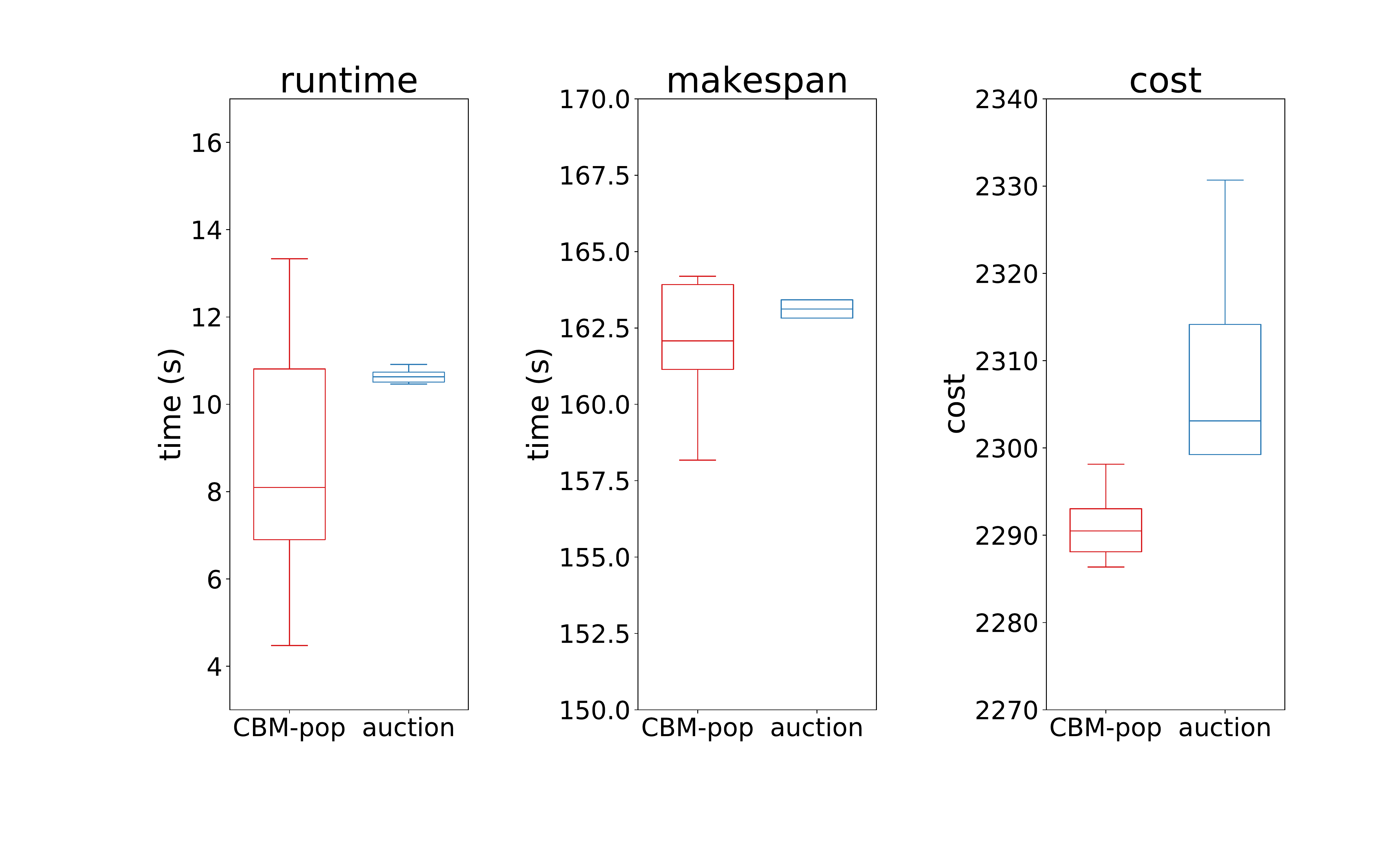}
    \caption{Comparison of the performance of the proposed CBM-pop (red) with the auction-based algorithm (blue). We observe the algorithm CPU time, solution makespan, and total cost for $10$ runs of each algorithm.}
    \label{fig:specularia_alg_comparison}
\end{figure}

We have run several simulations for the specified use case. As for the benchmark problem set, we compared the performance of the proposed algorithm with the auction-based state-of-the-art solution for task planning problems \cite{Nunes2017}. We ran $10$ simulations for each algorithm, and the results are consonant with the conclusions drawn earlier. In Figure \ref{fig:specularia_alg_comparison} we can observe CPU time, makespan, and cost distribution for the obtained solutions. For this example, the average reduction in CPU time of the proposed algorithm compared to the auction protocol is $13.23\%$. Regarding makespan and cost, a slight average improvement of $0.59\%$ and $0.76\%$, respectively, is observed.

For this relatively small problem, the qualities of the obtained solutions are very similar for both algorithms. However, we have previously shown that our algorithm adapts better to the problem complexity. Another important advantage is that our solution produces results in a distributed manner. The auction algorithm, on the other hand, assumes a central auctioneer agent that processes bids from other agents and makes task assignments. In distributed systems, this may be regarded as a tactically vulnerable point.

The final generated schedule for the SpECULARIA use case is displayed in \ref{appendix} for convenience. Tasks are color-coded according to the convention defined in Figure \ref{fig:greenhouse_layout}. Tasks of moving plants that do not require maintenance into the buffer are marked in gray. Precedence relations are indicated by arrows. Setup times are not displayed in this illustration to preserve a more compact representation of the schedule. Based on the generated schedule, we can conclude that the proposed method successfully handles the problem of planning tasks for this type of operations. The animated visualization of the execution of the obtained schedule is available at \cite{PaperWebpage}.

\section{Conclusion and Future Work}

In summary, in this paper we have developed a robust and fast task planning method for heterogeneous multi-robot systems. The planning method addresses problems with cross-schedule dependencies, in particular precedence constraints. We synthesized a general model that relates task planning (allocation and scheduling) to the well-studied VRP. This exposes task planning problems to various optimization techniques available in vehicle routing, which could lead to many compelling solutions for task planning in the future.

In this work, we have found a solution to the problem in a distributed manner by applying a metaheuristic approach based on evolutionary computation with knowledge sharing and mimetism. We have extensively tested the performance of the proposed algorithm. First, we ran simulations on an established set of benchmark examples of capacitated multi-depot VRP, where the algorithm was found to perform near-optimally, with a high computational speed. Next, we established a benchmark dataset repository for planning for tasks of class XD[ST-SR-TA] and tested the proposed algorithm against existing task planning methods. Simulation results show that the approach has better computational speed and scalability without loss of optimality compared to state-of-the-art distributed methods. We have also provided a novel application of the planning procedure to a real-world use case of a greenhouse maintained by a multi-robot system.

As future work, we are interested in testing the proposed approach in a more dynamic setting and introducing protocols for handling perturbations in the system, including asynchronous and stochastic arrival of new tasks in the system. Given the distributed nature of the proposed algorithm, we plan to consider robustness with respect to delays or information loss in the communication channel. Moreover, we aim to extend the modeling of task planning problems to the next level of complexity, the class Complex Dependency (CD), where each task can be achieved in multiple ways. The final unified system, incorporating the previously developed coordination framework and task planning for tasks of the highest complexity, promises compelling advances in task planning and will be the focus of our future efforts.

\section*{Acknowledgements}
This work has been supported by European Commission Horizon 2020 Programme through project under G. A. number 810321, named Twinning coordination action for spreading excellence in Aerial Robotics - AeRoTwin and by  Croatian Science Foundation under the project Specularia UIP-2017-05-4042. The work of doctoral student Barbara Arbanas has been supported in part by the “Young researchers’ career development project--training of doctoral students” of the Croatian Science Foundation funded by the European Union from the European Social Fund.

\balance
\bibliography{bibliography/els_main.bib}

\begin{thebibliography}{54}
\providecommand{\natexlab}[1]{#1}
\providecommand{\url}[1]{\texttt{#1}}
\providecommand{\urlprefix}{URL }
\expandafter\ifx\csname urlstyle\endcsname\relax
  \providecommand{\doi}[1]{doi:\discretionary{}{}{}#1}\else
  \providecommand{\doi}[1]{doi:\discretionary{}{}{}\begingroup
  \urlstyle{rm}\url{#1}\endgroup}\fi
\providecommand{\bibinfo}[2]{#2}

\bibitem[{Ismail and Sariff(2019)}]{Ismail2019}
\bibinfo{author}{Z.~H. Ismail}, \bibinfo{author}{N.~Sariff}, \bibinfo{title}{A
  Survey and Analysis of Cooperative Multi-Agent Robot Systems: Challenges and
  Directions}, in: \bibinfo{booktitle}{Applications of Mobile Robots},
  \bibinfo{publisher}{{IntechOpen}},
  \doi{\bibinfo{doi}{10.5772/intechopen.79337}}, \bibinfo{year}{2019}.

\bibitem[{Dahl et~al.(2009)Dahl, Matari{\'{c}}, and Sukhatme}]{Dahl2009}
\bibinfo{author}{T.~S. Dahl}, \bibinfo{author}{M.~Matari{\'{c}}},
  \bibinfo{author}{G.~S. Sukhatme}, \bibinfo{title}{Multi-robot task allocation
  through vacancy chain scheduling}, \bibinfo{journal}{Robotics and Autonomous
  Systems} \bibinfo{volume}{57}~(\bibinfo{number}{6-7}) (\bibinfo{year}{2009})
  \bibinfo{pages}{674--687}, \doi{\bibinfo{doi}{10.1016/j.robot.2008.12.001}}.

\bibitem[{Wawerla and Vaughan(2010)}]{Wawerla2010}
\bibinfo{author}{J.~Wawerla}, \bibinfo{author}{R.~T. Vaughan},
  \bibinfo{title}{A fast and frugal method for team-task allocation in a
  multi-robot transportation system}, in: \bibinfo{booktitle}{2010 {IEEE}
  International Conference on Robotics and Automation},
  \bibinfo{publisher}{{IEEE}}, \doi{\bibinfo{doi}{10.1109/robot.2010.5509865}},
  \bibinfo{year}{2010}.

\bibitem[{Bicchi et~al.(2008)Bicchi, Danesi, Dini, Porta, Pallottino, Savino,
  and Schiavi}]{Bicchi2008}
\bibinfo{author}{A.~Bicchi}, \bibinfo{author}{A.~Danesi},
  \bibinfo{author}{G.~Dini}, \bibinfo{author}{S.~Porta},
  \bibinfo{author}{L.~Pallottino}, \bibinfo{author}{I.~Savino},
  \bibinfo{author}{R.~Schiavi}, \bibinfo{title}{Heterogeneous Wireless
  Multirobot System}, \bibinfo{journal}{{IEEE} Robotics {\&} Automation
  Magazine} \bibinfo{volume}{15}~(\bibinfo{number}{1}) (\bibinfo{year}{2008})
  \bibinfo{pages}{62--70}, \doi{\bibinfo{doi}{10.1109/m-ra.2007.914925}}.

\bibitem[{Pei and Mutka(2012)}]{Pei2012}
\bibinfo{author}{Y.~Pei}, \bibinfo{author}{M.~W. Mutka},
  \bibinfo{title}{Steiner traveler: Relay deployment for remote sensing in
  heterogeneous multi-robot exploration}, in: \bibinfo{booktitle}{2012 {IEEE}
  International Conference on Robotics and Automation},
  \bibinfo{publisher}{{IEEE}}, \doi{\bibinfo{doi}{10.1109/icra.2012.6225347}},
  \bibinfo{year}{2012}.

\bibitem[{Zhang and Wong(2015)}]{Zhang2015}
\bibinfo{author}{L.~Zhang}, \bibinfo{author}{T.~Wong}, \bibinfo{title}{An
  object-coding genetic algorithm for integrated process planning and
  scheduling}, \bibinfo{journal}{European Journal of Operational Research}
  \bibinfo{volume}{244}~(\bibinfo{number}{2}) (\bibinfo{year}{2015})
  \bibinfo{pages}{434--444}, \doi{\bibinfo{doi}{10.1016/j.ejor.2015.01.032}}.

\bibitem[{Gombolay et~al.(2018)Gombolay, Wilcox, and Shah}]{Gombolay2018}
\bibinfo{author}{M.~Gombolay}, \bibinfo{author}{R.~Wilcox},
  \bibinfo{author}{J.~Shah}, \bibinfo{title}{{Fast Scheduling of Multi-Robot
  Teams with Temporospatial Constraints}}, \bibinfo{journal}{IEEE T-RO}
  (\bibinfo{year}{2018})
  \bibinfo{pages}{1--20}\doi{\bibinfo{doi}{10.15607/rss.2013.ix.049}}.

\bibitem[{Garc{\'{\i}}a et~al.(2013)Garc{\'{\i}}a, Caama{\~{n}}o, Duro, and
  Bellas}]{Garca2013}
\bibinfo{author}{P.~Garc{\'{\i}}a}, \bibinfo{author}{P.~Caama{\~{n}}o},
  \bibinfo{author}{R.~J. Duro}, \bibinfo{author}{F.~Bellas},
  \bibinfo{title}{Scalable Task Assignment for Heterogeneous Multi-Robot
  Teams}, \bibinfo{journal}{International Journal of Advanced Robotic Systems}
  \bibinfo{volume}{10}~(\bibinfo{number}{2}) (\bibinfo{year}{2013})
  \bibinfo{pages}{105}, \doi{\bibinfo{doi}{10.5772/55489}}.

\bibitem[{Capitan et~al.(2012)Capitan, Spaan, Merino, and Ollero}]{Capitan2012}
\bibinfo{author}{J.~Capitan}, \bibinfo{author}{M.~T. Spaan},
  \bibinfo{author}{L.~Merino}, \bibinfo{author}{A.~Ollero},
  \bibinfo{title}{Decentralized multi-robot cooperation with auctioned
  {POMDPs}}, in: \bibinfo{booktitle}{2012 {IEEE} International Conference on
  Robotics and Automation}, \bibinfo{publisher}{{IEEE}},
  \doi{\bibinfo{doi}{10.1109/icra.2012.6224917}}, \bibinfo{year}{2012}.

\bibitem[{Choi et~al.(2009)Choi, Brunet, and How}]{Choi2009}
\bibinfo{author}{H.-L. Choi}, \bibinfo{author}{L.~Brunet},
  \bibinfo{author}{J.~How}, \bibinfo{title}{Consensus-Based Decentralized
  Auctions for Robust Task Allocation}, \bibinfo{journal}{{IEEE} Transactions
  on Robotics} \bibinfo{volume}{25}~(\bibinfo{number}{4})
  (\bibinfo{year}{2009}) \bibinfo{pages}{912--926},
  \doi{\bibinfo{doi}{10.1109/tro.2009.2022423}}.

\bibitem[{Nunes et~al.(2017)Nunes, McIntire, and Gini}]{Nunes2017}
\bibinfo{author}{E.~Nunes}, \bibinfo{author}{M.~McIntire},
  \bibinfo{author}{M.~Gini}, \bibinfo{title}{Decentralized multi-robot
  allocation of tasks with temporal and precedence constraints},
  \bibinfo{journal}{Advanced Robotics}
  \bibinfo{volume}{31}~(\bibinfo{number}{22}) (\bibinfo{year}{2017})
  \bibinfo{pages}{1193--1207}, ISSN \bibinfo{issn}{0169-1864},
  \doi{\bibinfo{doi}{10.1080/01691864.2017.1396922}}.

\bibitem[{Korsah et~al.(2013)Korsah, Stentz, and Dias}]{korsahTaxonomy}
\bibinfo{author}{G.~A. Korsah}, \bibinfo{author}{A.~Stentz},
  \bibinfo{author}{M.~B. Dias}, \bibinfo{title}{A comprehensive taxonomy for
  multi-robot task allocation}, \bibinfo{journal}{The International Journal of
  Robotics Research} \bibinfo{volume}{32}~(\bibinfo{number}{12})
  (\bibinfo{year}{2013}) \bibinfo{pages}{1495--1512},
  \doi{\bibinfo{doi}{10.1177/0278364913496484}}.

\bibitem[{Mitiche et~al.(2019)Mitiche, Boughaci, and Gini}]{Mitiche2019}
\bibinfo{author}{H.~Mitiche}, \bibinfo{author}{D.~Boughaci},
  \bibinfo{author}{M.~Gini}, \bibinfo{title}{Iterated Local Search for
  Time-extended Multi-robot Task Allocation with Spatio-temporal and Capacity
  Constraints}, \bibinfo{journal}{Journal of Intelligent Systems}
  \bibinfo{volume}{28}~(\bibinfo{number}{2}) (\bibinfo{year}{2019})
  \bibinfo{pages}{347--360}, \doi{\bibinfo{doi}{10.1515/jisys-2018-0267}}.

\bibitem[{Saribatur et~al.(2014)Saribatur, Erdem, and Patoglu}]{Saribatur2014}
\bibinfo{author}{Z.~G. Saribatur}, \bibinfo{author}{E.~Erdem},
  \bibinfo{author}{V.~Patoglu}, \bibinfo{title}{Cognitive factories with
  multiple teams of heterogeneous robots: Hybrid reasoning for optimal feasible
  global plans}, in: \bibinfo{booktitle}{2014 {IEEE}/{RSJ} International
  Conference on Intelligent Robots and Systems}, \bibinfo{publisher}{{IEEE}},
  \doi{\bibinfo{doi}{10.1109/iros.2014.6942965}}, \bibinfo{year}{2014}.

\bibitem[{Schillinger et~al.(2018)Schillinger, Burger, and
  Dimarogonas}]{Schillinger2018}
\bibinfo{author}{P.~Schillinger}, \bibinfo{author}{M.~Burger},
  \bibinfo{author}{D.~V. Dimarogonas}, \bibinfo{title}{Auctioning over
  Probabilistic Options for Temporal Logic-Based Multi-Robot Cooperation Under
  Uncertainty}, in: \bibinfo{booktitle}{2018 {IEEE} International Conference on
  Robotics and Automation ({ICRA})}, \bibinfo{publisher}{{IEEE}},
  \doi{\bibinfo{doi}{10.1109/icra.2018.8462967}}, \bibinfo{year}{2018}.

\bibitem[{Omidshafiei et~al.(2017)Omidshafiei, Agha{\textendash}Mohammadi,
  Amato, Liu, How, and Vian}]{Omidshafiei2017}
\bibinfo{author}{S.~Omidshafiei},
  \bibinfo{author}{A.~Agha{\textendash}Mohammadi}, \bibinfo{author}{C.~Amato},
  \bibinfo{author}{S.~Liu}, \bibinfo{author}{J.~P. How},
  \bibinfo{author}{J.~Vian}, \bibinfo{title}{Decentralized control of
  multi-robot partially observable Markov decision processes using belief space
  macro-actions}, \bibinfo{journal}{The International Journal of Robotics
  Research} \bibinfo{volume}{36}~(\bibinfo{number}{2}) (\bibinfo{year}{2017})
  \bibinfo{pages}{231--258}, \doi{\bibinfo{doi}{10.1177/0278364917692864}}.

\bibitem[{Arbanas et~al.(2018)Arbanas, Ivanovic, Car, Orsag, Petrovic, and
  Bogdan}]{Arbanas2018}
\bibinfo{author}{B.~Arbanas}, \bibinfo{author}{A.~Ivanovic},
  \bibinfo{author}{M.~Car}, \bibinfo{author}{M.~Orsag},
  \bibinfo{author}{T.~Petrovic}, \bibinfo{author}{S.~Bogdan},
  \bibinfo{title}{Decentralized planning and control for {UAV}–{UGV}
  cooperative teams}, \bibinfo{journal}{Autonomous Robots}
  \doi{\bibinfo{doi}{10.1007/s10514-018-9712-y}}.

\bibitem[{Krizmancic et~al.(2020)Krizmancic, Arbanas, Petrovic, Petric, and
  Bogdan}]{Krizmancic2020}
\bibinfo{author}{M.~Krizmancic}, \bibinfo{author}{B.~Arbanas},
  \bibinfo{author}{T.~Petrovic}, \bibinfo{author}{F.~Petric},
  \bibinfo{author}{S.~Bogdan}, \bibinfo{title}{Cooperative Aerial-Ground
  Multi-Robot System for Automated Construction Tasks},
  \bibinfo{journal}{{IEEE} Robotics and Automation Letters}
  \bibinfo{volume}{5}~(\bibinfo{number}{2}) (\bibinfo{year}{2020})
  \bibinfo{pages}{798--805}, \doi{\bibinfo{doi}{10.1109/lra.2020.2965855}}.

\bibitem[{Korsah(2011)}]{Korsah2011}
\bibinfo{author}{G.~A. Korsah}, \bibinfo{title}{{Exploring Bounded Optimal
  Coordination for Heterogeneous Teams with Cross-Schedule Dependencies}}
  \doi{\bibinfo{doi}{10.1184/R1/6716585.v1}}.

\bibitem[{{Gurobi Optimization, LLC}(2019)}]{gurobi}
\bibinfo{author}{{Gurobi Optimization, LLC}}, \bibinfo{title}{Gurobi Optimizer
  Reference Manual}, \urlprefix\url{http://www.gurobi.com},
  \bibinfo{year}{2019}.

\bibitem[{Meignan et~al.(2010)Meignan, Koukam, and Créput}]{Meignan2009}
\bibinfo{author}{D.~Meignan}, \bibinfo{author}{A.~Koukam},
  \bibinfo{author}{J.-C. Créput}, \bibinfo{title}{Coalition-based
  metaheuristic: A self-adaptive metaheuristic using reinforcement learning and
  mimetism}, \bibinfo{journal}{Journal of Heuristics} \bibinfo{volume}{16}
  (\bibinfo{year}{2010}) \bibinfo{pages}{859--879},
  \doi{\bibinfo{doi}{10.1007/s10732-009-9121-7}}.

\bibitem[{Ben(2021)}]{BenchmarkRepo}
\bibinfo{title}{Task Planning Benchmark Repository},
  \bibinfo{howpublished}{\url{https://sites.google.com/view/taskplanningrepository/home}},
  \bibinfo{note}{accessed: 2021-02-19}, \bibinfo{year}{2021}.

\bibitem[{{Laboratory for Robotics {and} Intelligent Control
  Systems}(2017)}]{specularia}
\bibinfo{author}{{Laboratory for Robotics {and} Intelligent Control Systems}},
  \bibinfo{title}{Specularia},
  \bibinfo{howpublished}{\url{http://specularia.fer.hr}},
  \bibinfo{note}{accessed: 2021-02-19}, \bibinfo{year}{2017}.

\bibitem[{Garey and Johnson(1990)}]{Garey1990}
\bibinfo{author}{M.~R. Garey}, \bibinfo{author}{D.~S. Johnson},
  \bibinfo{title}{Computers and Intractability; A Guide to the Theory of
  NP-Completeness}, \bibinfo{publisher}{W. H. Freeman \& Co.},
  \bibinfo{address}{USA}, ISBN \bibinfo{isbn}{0716710455},
  \bibinfo{year}{1990}.

\bibitem[{Laporte et~al.(1984)Laporte, Nobert, and D}]{Laporte1984}
\bibinfo{author}{G.~Laporte}, \bibinfo{author}{Y.~Nobert},
  \bibinfo{author}{A.~D}, \bibinfo{title}{{Optimal solutions to capacitated
  multidepot vehicle routing problems}}, \bibinfo{journal}{Congressus
  Numerantium} \bibinfo{volume}{44} (\bibinfo{year}{1984})
  \bibinfo{pages}{283--292}.

\bibitem[{Laporte et~al.(1988)Laporte, Norbert, and Taillefer}]{Laporte1988}
\bibinfo{author}{G.~Laporte}, \bibinfo{author}{Y.~Norbert},
  \bibinfo{author}{S.~Taillefer}, \bibinfo{title}{Solving a Family of
  Multi-Depot Vehicle Routing and Location-Routing Problems},
  \bibinfo{journal}{Transportation Science}
  \bibinfo{volume}{22}~(\bibinfo{number}{3}) (\bibinfo{year}{1988})
  \bibinfo{pages}{161--172}, ISSN \bibinfo{issn}{00411655, 15265447}.

\bibitem[{Baldacci and Mingozzi(2009)}]{Baldacci2009}
\bibinfo{author}{R.~Baldacci}, \bibinfo{author}{A.~Mingozzi},
  \bibinfo{title}{An unified exact method for solving different classes of
  vehicle routing problems}, \bibinfo{journal}{Mathematical Programming}
  \bibinfo{volume}{120} (\bibinfo{year}{2009}) \bibinfo{pages}{347--380},
  \doi{\bibinfo{doi}{10.1007/s10107-008-0218-9}}.

\bibitem[{Contardo and Martinelli(2014)}]{Contardo2014}
\bibinfo{author}{C.~Contardo}, \bibinfo{author}{R.~Martinelli},
  \bibinfo{title}{A new exact algorithm for the multi-depot vehicle routing
  problem under capacity and route length constraints},
  \bibinfo{journal}{Discrete Optimization} \bibinfo{volume}{12}
  (\bibinfo{year}{2014}) \bibinfo{pages}{129 -- 146}, ISSN
  \bibinfo{issn}{1572-5286}, \doi{\bibinfo{doi}{10.1016/j.disopt.2014.03.001}}.

\bibitem[{Montoya-Torres et~al.(2015)Montoya-Torres, {López Franco}, {Nieto
  Isaza}, {Felizzola Jiménez}, and Herazo-Padilla}]{MontoyaTorres2015}
\bibinfo{author}{J.~R. Montoya-Torres}, \bibinfo{author}{J.~{López Franco}},
  \bibinfo{author}{S.~{Nieto Isaza}}, \bibinfo{author}{H.~{Felizzola
  Jiménez}}, \bibinfo{author}{N.~Herazo-Padilla}, \bibinfo{title}{A literature
  review on the vehicle routing problem with multiple depots},
  \bibinfo{journal}{Computers \& Industrial Engineering} \bibinfo{volume}{79}
  (\bibinfo{year}{2015}) \bibinfo{pages}{115--129}, ISSN
  \bibinfo{issn}{0360-8352}, \doi{\bibinfo{doi}{10.1016/j.cie.2014.10.029}}.

\bibitem[{Abdel-Basset et~al.(2018)Abdel-Basset, Abdel-Fatah, and
  Sangaiah}]{Basset2018}
\bibinfo{author}{M.~Abdel-Basset}, \bibinfo{author}{L.~Abdel-Fatah},
  \bibinfo{author}{A.~K. Sangaiah}, \bibinfo{title}{Chapter 10 - Metaheuristic
  Algorithms: A Comprehensive Review}, in: \bibinfo{editor}{A.~K. Sangaiah},
  \bibinfo{editor}{M.~Sheng}, \bibinfo{editor}{Z.~Zhang} (Eds.),
  \bibinfo{booktitle}{Computational Intelligence for Multimedia Big Data on the
  Cloud with Engineering Applications}, Intelligent Data-Centric Systems,
  \bibinfo{publisher}{Academic Press}, ISBN \bibinfo{isbn}{978-0-12-813314-9},
  \bibinfo{pages}{185 -- 231},
  \doi{\bibinfo{doi}{10.1016/B978-0-12-813314-9.00010-4}},
  \bibinfo{year}{2018}.

\bibitem[{Ombuki-Berman and Hanshar(2008)}]{Berman2008}
\bibinfo{author}{B.~Ombuki-Berman}, \bibinfo{author}{F.~Hanshar},
  \bibinfo{title}{Using Genetic Algorithms for Multi-depot Vehicle Routing},
  vol. \bibinfo{volume}{161}, ISBN \bibinfo{isbn}{978-3-540-85151-6},
  \bibinfo{pages}{77--99}, \doi{\bibinfo{doi}{10.1007/978-3-540-85152-3\_4}},
  \bibinfo{year}{2008}.

\bibitem[{Gesú et~al.(2005)Gesú, Giancarlo, Lo~Bosco, Raimondi, and
  Scaturro}]{Gesu2005}
\bibinfo{author}{V.~Gesú}, \bibinfo{author}{R.~Giancarlo},
  \bibinfo{author}{G.~Lo~Bosco}, \bibinfo{author}{A.~Raimondi},
  \bibinfo{author}{D.~Scaturro}, \bibinfo{title}{GenClust: A Genetic Algorithm
  for Clustering Gene Expression Data}, \bibinfo{journal}{BMC bioinformatics}
  \bibinfo{volume}{6} (\bibinfo{year}{2005}) \bibinfo{pages}{289},
  \doi{\bibinfo{doi}{10.1186/1471-2105-6-289}}.

\bibitem[{{Mahmud} and {Haque}(2019)}]{Mahmud2019}
\bibinfo{author}{N.~{Mahmud}}, \bibinfo{author}{M.~M. {Haque}},
  \bibinfo{title}{Solving Multiple Depot Vehicle Routing Problem (MDVRP) using
  Genetic Algorithm}, in: \bibinfo{booktitle}{2019 International Conference on
  Electrical, Computer and Communication Engineering (ECCE)},
  \bibinfo{pages}{1--6}, \doi{\bibinfo{doi}{10.1109/ECACE.2019.8679429}},
  \bibinfo{year}{2019}.

\bibitem[{{Venkata Narasimha} et~al.(2013){Venkata Narasimha}, Kivelevitch,
  Sharma, and Kumar}]{Narasimha2013}
\bibinfo{author}{K.~{Venkata Narasimha}}, \bibinfo{author}{E.~Kivelevitch},
  \bibinfo{author}{B.~Sharma}, \bibinfo{author}{M.~Kumar}, \bibinfo{title}{An
  ant colony optimization technique for solving min–max Multi-Depot Vehicle
  Routing Problem}, \bibinfo{journal}{Swarm and Evolutionary Computation}
  \bibinfo{volume}{13} (\bibinfo{year}{2013}) \bibinfo{pages}{63 -- 73}, ISSN
  \bibinfo{issn}{2210-6502}, \doi{\bibinfo{doi}{10.1016/j.swevo.2013.05.005}}.

\bibitem[{Zhang et~al.(2019)Zhang, Zhang, Gajpal, and Appadoo}]{Zhang2019}
\bibinfo{author}{S.~Zhang}, \bibinfo{author}{W.~Zhang},
  \bibinfo{author}{Y.~Gajpal}, \bibinfo{author}{S.~S. Appadoo},
  \bibinfo{title}{Ant Colony Algorithm for Routing Alternate Fuel Vehicles in
  Multi-depot Vehicle Routing Problem}, \bibinfo{publisher}{Springer
  Singapore}, ISBN \bibinfo{isbn}{978-981-13-0860-4},
  \bibinfo{pages}{251--260}, \doi{\bibinfo{doi}{10.1007/978-981-13-0860-4_19}},
  \bibinfo{year}{2019}.

\bibitem[{Escobar et~al.(2014)Escobar, Linfati, Toth, and
  Baldoquin}]{Escobar2014}
\bibinfo{author}{J.~Escobar}, \bibinfo{author}{R.~Linfati},
  \bibinfo{author}{P.~Toth}, \bibinfo{author}{M.~Baldoquin}, \bibinfo{title}{A
  hybrid Granular Tabu Search algorithm for the Multi-Depot Vehicle Routing
  Problem}, \bibinfo{journal}{Journal of Heuristics} \bibinfo{volume}{20}
  (\bibinfo{year}{2014}) \bibinfo{pages}{1--27},
  \doi{\bibinfo{doi}{10.1007/s10732-014-9247-0}}.

\bibitem[{Luo and Chen(2014)}]{Luo2014}
\bibinfo{author}{J.~Luo}, \bibinfo{author}{M.-R. Chen},
  \bibinfo{title}{Multi-phase modified shuffled frog leaping algorithm with
  extremal optimization for the {MDVRP} and the {MDVRPTW}},
  \bibinfo{journal}{Computers \& Industrial Engineering} \bibinfo{volume}{72}
  (\bibinfo{year}{2014}) \bibinfo{pages}{84 -- 97}, ISSN
  \bibinfo{issn}{0360-8352}, \doi{\bibinfo{doi}{10.1016/j.cie.2014.03.004}}.

\bibitem[{Mirabi et~al.(2010)Mirabi, {Fatemi Ghomi}, and Jolai}]{Mirabi2010}
\bibinfo{author}{M.~Mirabi}, \bibinfo{author}{S.~{Fatemi Ghomi}},
  \bibinfo{author}{F.~Jolai}, \bibinfo{title}{Efficient stochastic hybrid
  heuristics for the multi-depot vehicle routing problem},
  \bibinfo{journal}{Robotics and Computer-Integrated Manufacturing}
  \bibinfo{volume}{26}~(\bibinfo{number}{6}) (\bibinfo{year}{2010})
  \bibinfo{pages}{564 -- 569}, ISSN \bibinfo{issn}{0736-5845},
  \doi{\bibinfo{doi}{10.1016/j.rcim.2010.06.023}}, \bibinfo{note}{19th
  International Conference on Flexible Automation and Intelligent
  Manufacturing}.

\bibitem[{{Haerani} et~al.(2017){Haerani}, {Wardhani}, {Putri}, and
  {Sukmana}}]{Haerani2017}
\bibinfo{author}{E.~{Haerani}}, \bibinfo{author}{L.~K. {Wardhani}},
  \bibinfo{author}{D.~K. {Putri}}, \bibinfo{author}{H.~T. {Sukmana}},
  \bibinfo{title}{Optimization of multiple depot vehicle routing problem
  (MDVRP) on perishable product distribution by using genetic algorithm and
  fuzzy logic controller (FLC)}, in: \bibinfo{booktitle}{2017 5th International
  Conference on Cyber and IT Service Management (CITSM)},
  \bibinfo{pages}{1--5}, \doi{\bibinfo{doi}{10.1109/CITSM.2017.8089314}},
  \bibinfo{year}{2017}.

\bibitem[{Stodola(2018)}]{Stodola2018}
\bibinfo{author}{P.~Stodola}, \bibinfo{title}{Using Metaheuristics on the
  Multi-Depot Vehicle Routing Problem with Modified Optimization Criterion},
  \bibinfo{journal}{Algorithms} \bibinfo{volume}{11}~(\bibinfo{number}{5}),
  ISSN \bibinfo{issn}{1999-4893}, \doi{\bibinfo{doi}{10.3390/a11050074}}.

\bibitem[{{de Oliveira} et~al.(2016){de Oliveira}, Enayatifar, Sadaei,
  Guimarães, and Potvin}]{deOliveira2016}
\bibinfo{author}{F.~B. {de Oliveira}}, \bibinfo{author}{R.~Enayatifar},
  \bibinfo{author}{H.~J. Sadaei}, \bibinfo{author}{F.~G. Guimarães},
  \bibinfo{author}{J.-Y. Potvin}, \bibinfo{title}{A cooperative coevolutionary
  algorithm for the Multi-Depot Vehicle Routing Problem},
  \bibinfo{journal}{Expert Systems with Applications} \bibinfo{volume}{43}
  (\bibinfo{year}{2016}) \bibinfo{pages}{117--130}, ISSN
  \bibinfo{issn}{0957-4174}, \doi{\bibinfo{doi}{10.1016/j.eswa.2015.08.030}}.

\bibitem[{Sadati et~al.(2020)Sadati, Aksen, and Aras}]{HesamSadati2020}
\bibinfo{author}{M.~E.~H. Sadati}, \bibinfo{author}{D.~Aksen},
  \bibinfo{author}{N.~Aras}, \bibinfo{title}{The r-interdiction selective
  multi-depot vehicle routing problem}, \bibinfo{journal}{International
  Transactions in Operational Research}
  \bibinfo{volume}{27}~(\bibinfo{number}{2}) (\bibinfo{year}{2020})
  \bibinfo{pages}{835--866}, \doi{\bibinfo{doi}{10.1111/itor.12669}}.

\bibitem[{Chang(2015)}]{CHANG2015}
\bibinfo{author}{K.-H. Chang}, \bibinfo{title}{Chapter 19 - Multiobjective
  Optimization and Advanced Topics}, in: \bibinfo{editor}{K.-H. Chang} (Ed.),
  \bibinfo{booktitle}{e-Design}, \bibinfo{publisher}{Academic Press},
  \bibinfo{address}{Boston}, ISBN \bibinfo{isbn}{978-0-12-382038-9},
  \bibinfo{pages}{1105 -- 1173},
  \doi{\bibinfo{doi}{10.1016/B978-0-12-382038-9.00019-3}},
  \bibinfo{year}{2015}.

\bibitem[{Salhi et~al.(2014)Salhi, Imran, and Wassan}]{Salhi2014}
\bibinfo{author}{S.~Salhi}, \bibinfo{author}{A.~Imran}, \bibinfo{author}{N.~A.
  Wassan}, \bibinfo{title}{The multi-depot vehicle routing problem with
  heterogeneous vehicle fleet: Formulation and a variable neighborhood search
  implementation}, \bibinfo{journal}{Computers \& Operations Research}
  \bibinfo{volume}{52} (\bibinfo{year}{2014}) \bibinfo{pages}{315 -- 325}, ISSN
  \bibinfo{issn}{0305-0548}, \doi{\bibinfo{doi}{10.1016/j.cor.2013.05.011}},
  \bibinfo{note}{recent advances in Variable neighborhood search}.

\bibitem[{Toth et~al.(2015)Toth, Vigo, for Industrial, and
  Mathematics}]{toth2015VRP}
\bibinfo{author}{P.~Toth}, \bibinfo{author}{D.~Vigo}, \bibinfo{author}{S.~for
  Industrial}, \bibinfo{author}{A.~Mathematics}, \bibinfo{title}{Vehicle
  Routing: Problems, Methods, and Applications}, {MOS}-{SIAM} series on
  optimization, \bibinfo{publisher}{Society for Industrial and Applied
  Mathematics (SIAM, 3600 Market Street, Floor 6, Philadelphia, PA 19104)},
  ISBN \bibinfo{isbn}{9781523109371},
  \urlprefix\url{https://books.google.hr/books?id=YL\_CswEACAAJ},
  \bibinfo{year}{2015}.

\bibitem[{Li et~al.(2007)Li, Golden, and Wasil}]{Li2007openVRP}
\bibinfo{author}{F.~Li}, \bibinfo{author}{B.~Golden},
  \bibinfo{author}{E.~Wasil}, \bibinfo{title}{The open vehicle routing problem:
  Algorithms, large-scale test problems, and computational results},
  \bibinfo{journal}{Computers \& Operations Research}
  \bibinfo{volume}{34}~(\bibinfo{number}{10}) (\bibinfo{year}{2007})
  \bibinfo{pages}{2918 -- 2930}, ISSN \bibinfo{issn}{0305-0548},
  \doi{\bibinfo{doi}{10.1016/j.cor.2005.11.018}}.

\bibitem[{{Kennedy} and {Eberhart}(1995)}]{Kennedy95}
\bibinfo{author}{J.~{Kennedy}}, \bibinfo{author}{R.~{Eberhart}},
  \bibinfo{title}{Particle swarm optimization}, in:
  \bibinfo{booktitle}{Proceedings of ICNN'95 - International Conference on
  Neural Networks}, vol.~\bibinfo{volume}{4}, \bibinfo{pages}{1942--1948
  vol.4}, \doi{\bibinfo{doi}{10.1109/ICNN.1995.488968}}, \bibinfo{year}{1995}.

\bibitem[{Sprecher et~al.(1995)Sprecher, Kolisch, and Drexl}]{Sprecher1995}
\bibinfo{author}{A.~Sprecher}, \bibinfo{author}{R.~Kolisch},
  \bibinfo{author}{A.~Drexl}, \bibinfo{title}{Semi-active, active, and
  non-delay schedules for the resource-constrained project scheduling problem},
  \bibinfo{journal}{European Journal of Operational Research}
  \bibinfo{volume}{80}~(\bibinfo{number}{1}) (\bibinfo{year}{1995})
  \bibinfo{pages}{94--102}, ISSN \bibinfo{issn}{0377-2217},
  \doi{\bibinfo{doi}{10.1016/0377-2217(93)E0294-8}}.

\bibitem[{Deb and Kalyanmoy(2001)}]{Kalyanmoy2001}
\bibinfo{author}{K.~Deb}, \bibinfo{author}{D.~Kalyanmoy},
  \bibinfo{title}{Multi-Objective Optimization Using Evolutionary Algorithms},
  \bibinfo{publisher}{John Wiley \& Sons, Inc.}, \bibinfo{address}{USA}, ISBN
  \bibinfo{isbn}{047187339X}, \bibinfo{year}{2001}.

\bibitem[{Baptista~Pereira and Tavares(2009)}]{Pereira2009}
\bibinfo{editor}{F.~Baptista~Pereira}, \bibinfo{editor}{J.~Tavares} (Eds.),
  \bibinfo{title}{Bio-inspired Algorithms for the Vehicle Routing Problem},
  vol. \bibinfo{volume}{161}, \bibinfo{publisher}{Springer-Verlag Berlin
  Heidelberg}, \bibinfo{edition}{1} edn., ISBN
  \bibinfo{isbn}{978-3-540-85152-3},
  \doi{\bibinfo{doi}{10.1007/978-3-540-85152-3}}, \bibinfo{year}{2009}.

\bibitem[{{Stanford Artificial Intelligence Laboratory et al.}(2018)}]{ros}
\bibinfo{author}{{Stanford Artificial Intelligence Laboratory et al.}},
  \bibinfo{title}{Robotic Operating System},
  \urlprefix\url{https://www.ros.org}, \bibinfo{year}{2018}.

\bibitem[{E.T.S.I.~Informática(2013)}]{CordeauDataset}
\bibinfo{author}{U.~o.~M. E.T.S.I.~Informática}, \bibinfo{title}{Cordeau
  Multiple Depot VRP Instances},
  \bibinfo{howpublished}{\url{https://neo.lcc.uma.es/vrp/vrp-instances/multiple-depot-vrp-instances/}},
  \bibinfo{note}{accessed: 2021-02-19}, \bibinfo{year}{2013}.

\bibitem[{Cordeau et~al.(1997)Cordeau, Gendreau, and Laporte}]{Cordeau97}
\bibinfo{author}{J.-F. Cordeau}, \bibinfo{author}{M.~Gendreau},
  \bibinfo{author}{G.~Laporte}, \bibinfo{title}{A tabu search heuristic for
  periodic and multi-depot vehicle routing problems},
  \bibinfo{journal}{Networks} \bibinfo{volume}{30}~(\bibinfo{number}{2})
  (\bibinfo{year}{1997}) \bibinfo{pages}{105--119},
  \doi{\bibinfo{doi}{10.1002/(SICI)1097-0037(199709)30:2<105::AID-NET5>3.0.CO;2-G}}.

\bibitem[{Pap(2021)}]{PaperWebpage}
\bibinfo{title}{Distributed Allocation and Scheduling of Tasks With
  Cross-schedule Dependencies for Heterogeneous Multi-robot Teams},
  \bibinfo{howpublished}{\url{https://sites.google.com/view/vrp-task-planning/home}},
  \bibinfo{note}{accessed: 2021-02-19}, \bibinfo{year}{2021}.

\end{thebibliography}

\clearpage
\appendix
\section{SpECULARIA use case schedule}
\label{appendix}

\noindent\begin{minipage}{\textwidth}
    \includegraphics[angle=90,scale=0.55]{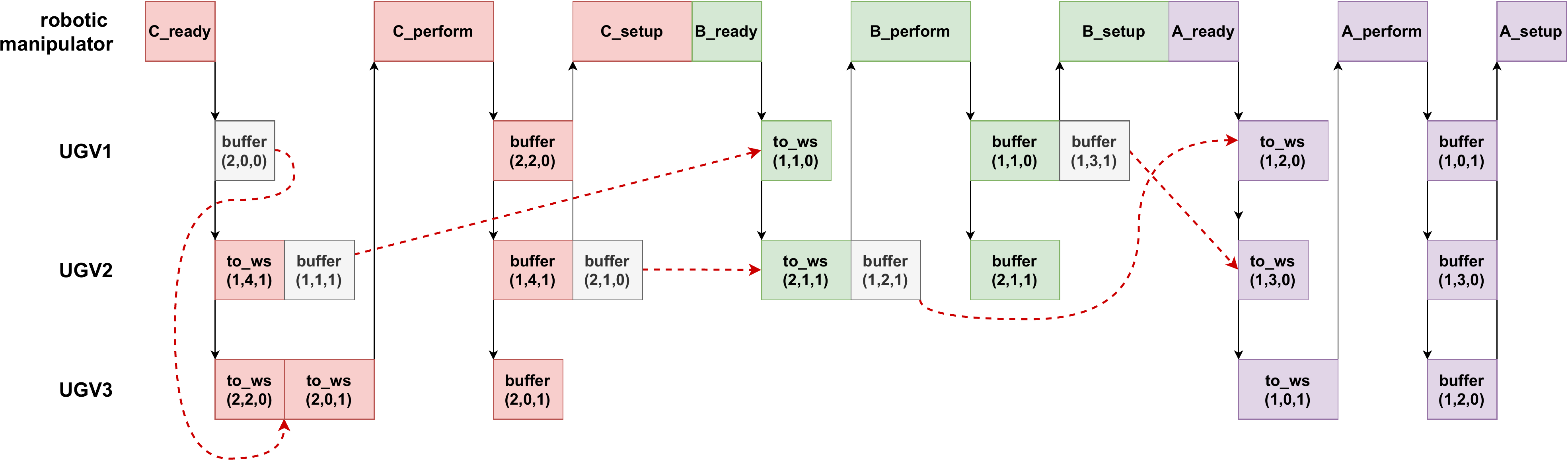}
    \centering
    \label{fig:specularia_schedule}
\end{minipage}

\end{document}